\documentclass[iicol]{resources/sn-jnl}
 
\usepackage{amsmath,amssymb,theorem}
\usepackage{graphicx}
\usepackage{algorithm,algpseudocodex}

\usepackage{color,xspace}

\usepackage{siunitx}

\usepackage{caption}
\usepackage{subcaption}

\usepackage{natbib}

\newtheorem{theorem}{Theorem}[section]

\newtheorem{remark}[theorem]{Remark}

\newtheorem{problem}[theorem]{Problem}
\newtheorem{assumption}[theorem]{Assumption}

\newcommand{\real}{{\mathbb{R}}}

\newcommand{\realnonnegative}{\mathbb{R}_{\ge 0}}
\newcommand{\integernonnegative}{\mathbb{Z}_{\ge 0}}

\newcommand{\diag}[1]{\operatorname{diag}\left( #1\right)}

\renewcommand{\tilde}{\widetilde}

\newcommand{\vmax}{v_{\text{max}}}
\newcommand{\timestep}{\Delta t}

\renewcommand{\epsilon}{\varepsilon}

\DeclareMathOperator*{\argmin}{\operatorname{arg\,min}}

\newcommand{\algostep}[1]{{\small\texttt{#1:}}\xspace}

\newcommand{\TwoNorm}[1]{\|#1\|}

\renewcommand{\hat}{\widehat}

\newcommand{\oprocendsymbol}{\hbox{$\bullet$}}
\newcommand{\oprocend}{\relax\ifmmode\else\unskip\hfill\fi\oprocendsymbol}

\renewcommand{\theenumi}{(\roman{enumi})}

\newcommand{\comment}[1]{}

\newcommand{\robots}{\mathcal{R}_s}
\newcommand{\robotsn}{N_s}
\newcommand{\robotpos}{p_i}
\newcommand{\robotvel}{v_i}
\newcommand{\robotdynams}{F_i}

\newcommand{\robotinput}{u_i}
\newcommand{\robotwayptabs}{p_i^*}

\newcommand{\targets}{\mathcal{R}_h}
\newcommand{\targetsn}{N_h}
\newcommand{\targetstate}{x_k}
\newcommand{\targetstaterel}{x_{i, k}}

\newcommand{\targetprcsnoise}{w_k}
\newcommand{\targetprcscov}{Q_k}
\newcommand{\targetprcscovmax}{\bar{Q}_k}
\newcommand{\targetdetection}{\mathbf{D}_k}
\newcommand{\targetmeasnoise}{v_{i, k}}
\newcommand{\targetsensraw}{\hat{x}_{i, k}^{\text{raw}}}
\newcommand{\targetsensrawmv}[1]{\hat{x}_{{#1}, k}^{\text{raw}}}
\newcommand{\targetsenspos}{\hat{x}_{i, k}}
\newcommand{\targetsensposmv}[1]{\hat{x}_{{#1}, k}}
\newcommand{\targetsenscov}{\Sigma_{i, k}}
\newcommand{\targetsenscovmap}{R_i^s}
\newcommand{\targetsenscomb}{\hat{x}_{i, k}^{c}}

\newcommand{\robotfov}{\operatorname{FOV}_i}
\newcommand{\robotfovshrink}{\Upsilon _{\text{sh}}}

\newcommand{\targetentropylocal}{\mathbf{h}(\targetsenspos |\targetsensraw(0:t))}
\newcommand{\targetentropylocalnv}[1]{\mathbf{h}(\targetsensposmv{#1} |\targetsensrawmv{#1}(0:t))}

\newcommand{\neighposrec}{\hat{p}_{i, j}^+}
\newcommand{\neighpos}{\hat{p}_{i, j}}
\newcommand{\neighposcov}{\Sigma_{i, j}}
\newcommand{\neighposcovmap}{R_i^p}

\newcommand{\robotposlast}{\Delta p_i}
\newcommand{\robotposlastcov}{\Sigma_i^{\Delta p}}
\newcommand{\robotposlastcovmap}{R_i^{\Delta p}}

\newcommand{\commrange}{r_c}

\newcommand{\costtracking}{\mathcal{H}}

\newcommand{\senslistlocal}{L_i^s}
\newcommand{\senslistrec}{L_{i, j}^{s+}}
\newcommand{\senslistneig}{L_{i, j}^{s}}
\newcommand{\targetsensneighpos}{\hat{x}_{i, j, k}}
\newcommand{\targetsensneighcov}{\Sigma_{i, j, k}}
\newcommand{\targetcovmax}{\bar{\Sigma}_i}
\newcommand{\targetchosen}{k_i^*} 
\newcommand{\targetsensneighlocalpos}{\hat{x}_{i, k}^j}

\newcommand{\phercov}{\Sigma_{i, s}^p}

\newcommand{\pherweightinit}{w_i^{\text{init}}}
\newcommand{\pherweightdecay}{w_i^\delta}
\newcommand{\pherweightthresh}{w_i^\diamond }

\newcommand{\pherneighcov}{\Sigma_{i, j, s}^p}

\begin{document}
    \title{Real-Time Distributed Infrastructure-free Searching and Target Tracking via Virtual Pheromones}

    \author*[1]{\fnm{Joseph} \sur{Prince Mathew}}\email{jprincem@gmu.edu}
    \author[1]{\fnm{Cameron} \sur{Nowzari}}\email{cnowzari@gmu.edu}
    \affil*[1]{\orgname{George Mason University}, \city{Fairfax}, \state{VA}-\postcode{22030}, \country{USA} }

    \abstract{
        Actively searching for targets using a multi-agent system in an unknown environment poses a two-pronged problem. On the one hand we need agents to cover as much of the environment as possible with little overlap and on the other hand the agents must coordinate among themselves to select and track targets thereby improving detection performance. This paper proposes a fully distributed solution for an ad-hoc network of agents to cooperatively search for targets and monitor them in an unknown infrastructure-free environment. The solution combines a distributed pheromone-based coverage control strategy with a distributed target selection mechanism. We show the effectiveness of the proposed solution using Monte-Carlo simulations and then validate the algorithm in a robotic system consisting of Lighter Than Air (LTA) agents.
    }

    \maketitle

    %%%%%%%%%%%%%%%%%%%%%%%%%%%%%%%%%%%%%%%%%%%%%%%%%%%%%%%%%%%%%%%%%%%%%%%%%%%%%%%%%%%%%%%%%%%%%%%%%%%%%%%%%%%%%%%%%%%%
    % Introduction
    %%%%%%%%%%%%%%%%%%%%%%%%%%%%%%%%%%%%%%%%%%%%%%%%%%%%%%%%%%%%%%%%%%%%%%%%%%%%%%%%%%%%%%%%%%%%%%%%%%%%%%%%%%%%%%%%%%%%
    \section{Introduction}
    \label{sec:Introduction}
        Searching an environment and tracking a priori unknown targets, a.k.a. active perception, forms one of the cornerstone tasks of many robotic applications. These include: search and rescue \cite{Queralta2020}, environmental process monitoring \cite{Liu2020, Villarrubia2017}, and traversing hazardous environments. For searching and tracking, utilizing multiple mobile sensors can reduce the time taken to cover a large area.Existing work shows both centralized \cite{Atanasov2015,Schlotfeldt2018,Kantaros2021} and decentralized \cite{Chen2021, Zhong2011} approaches to multi-robot information gathering. Decentralized solutions are preferred for systems requiring scalability and reliability. 
        
        Consider a scenario where a team of robotic agents must search an environment for multiple targets while wireless communicating. In cases where prior information about the environment is available, and/or global positioning of agents is enabled (e.g., GPS \cite{Chen2021}, or overhead cameras/VICON \cite{Viseras2020}), there is an ample amount of research dedicating to optimally searching such spaces. The closest area of research related to this type of active perception problem utilizes tree-based offline planning; where for a given planning horizon and initial configuration of the mobile sensors, all possible configurations for the subsequent time-steps are computed~\cite{Atanasov2015, Schlotfeldt2018, Kantaros2021}. One of the main requirements is to have all robots use information from every other robot, either as a centralized collation of sensor data \cite{Atanasov2015,Kantaros2021}, or connected graph in a distributed version \cite{Schlotfeldt2018}. Since these policies are computed offline, the deployment of these algorithms requires prior knowledge of the environment and hence it's very difficult to transfer the plan to a new environment. Reinforcement learning-based solutions have also been considered which also require prior knowledge of the environment to be searched~\cite{Jeong2019,Viseras2019,Best2018}. Further, there are POMDP based search and tracking strategies~\cite{Hollinger2009, Nguyen2019}, however, these algorithms do not scale well with growing environment size as well as number of agents and targets. They also require the robots to know the global positions for the formulation to work. The strong GPS requirement limits the deployment of such solutions in many environments, like, GPS-denied environments or an environment inaccessible to set up infrastructure. To relax this constraint we can employ relative positioning systems~\cite{Chen2022}, where we have seen RF solutions utilizing WiFi\cite{Retscher2020, Tong2021}, Bluetooth\cite{Wang2016, Obreja2020, Baronti2018} and UWB systems\cite{Tiemann2020, Heydariaan2020, Macoir2019, Silva2014} or vision based solutions like RGBD Camera \cite{Zhang2020, Pavliv2021} or LIDAR based SLAM systems \cite{Vodisch2022, Cho2022}.

        \subsection{Limited-View Searching and Tracking without Prior Information}
        
        Let us now consider a much harder version of this problem where no prior information is available, including the number of potential targets to be found. Additionally, consider that the robots have don't have access to their global positions nor the span/map of the environment. Designing optimal strategies for searching and tracking problem in such cases is intractable due to unavailability of information. For instance, robots will not know where they have searched in the environment, nor do they know how many targets are left to be tracked in the environment. Thus, common strategies in these scenarios involve simply randomly searching environments and finding targets~\cite{Robin2016, Cao2006, Sutantyo2010}. Other search strategies utilize specific patterns for search like lawnmower~\cite{Azpúrua2018, Alpern2003} pattern, spiral~\cite{WangMeghjani2022, Cabreira2018} pattern or other efficient search paths depending on the application.
      
        Although the different pieces of the solution to distributed multi-agent search and track exists, extending those to the limited information version while maintaining properties of the various sub-algorithms of the problem is non-trivial. Can these optimal solutions, whose optimality depends on information that may be unavailable or unverifiable in the real-world scenario, be somehow modified to be effective in the limited information case? In this paper we explore this possibility using a systems' integration approach. To that end, we partition our searching and tracking problem into two sub-problems which have clearly related works in the literature we can borrow from: \textbf{dynamic coverage control} and \textbf{dynamic target assignment}. We take the basic theory from these algorithms and modify them to be applicable with real-world implementations and demonstrate our results on real hardware as a proof of concept.  

        For dynamic coverage control, sophisticated strategies like Voronoi-based methods~\cite{Nowzari2012,Arslan2016, Luo2019, Chen2021} could be used, but, these require agents with complex sensing and communication capabilities. For example, they require communicating to agents at arbitrary long distance to/from their neighbors, or they need the ability to sense the target density function. These requirements may not be feasible in a priori unknown environment or supported by current hardware. The closest applicable solution is a probability-based map to determine previously explored areas and prioritize unexplored areas. This has been known as awareness control~\cite{Wang2010, Zhong2011}, Anti-Flocking~\cite{Ganganath2016, Wang2021} or Pheromone-based coverage control~\cite{Wang2022}, where the explored areas get higher scalar weights and the controller points the mobile sensor towards areas with lower weights. This strategy has been shown to work with multi-agent systems, however, we see that these systems require a central pheromone processing system \cite{Filipescu2009, Bottone2016, Oliveira2014}. They also include assumptions that all robots communicate with each other (thereby having instant access to the other's data) or use pre-existing infrastructure like RFID tags or light based region sensing \cite{Sun2019, Sugawara2004} or access to global agent positions~\cite{Ganganath2016}. These requirements limit the use of these implementations in large unknown environments. The decentralized storage and propagation of pheromones effectively in the multi-agent system is an area that require more investigation and our work with the virtual pheromone is directly aimed at this. 

        For dynamic target assignment, we look to linear sum assignment problems where two of the most popular solutions are Hungarian algorithm~\cite{Crouse2016, Chopra2017} and Auction algorithms~\cite{bertsekas1979, Zavlanos2008}. We have seen centralized target assignment solutions in \cite{Kantaros2021, Qie2019, Zhou2011}, where the centralized processing system has all the information and can easily make target assignment decisions as all information required to make the decision is readily available. This becomes nontrivial in a decentralized implementation of active perception due to the inherent asymmetry in the data available to each agent as they move in the environment. To overcome this, we have observed other works that  have the agents make sequentially distributed decisions as seen in \cite{Schlotfeldt2018, Volle2016}. This, however, does not scale well with large number of agents. One way to ensure optimal target assignment is to ensure connectivity between agents~\cite{Xia2022, zahroof2022}, however, connectivity maintenance works against exploration required for searching. Additionally, the agents may not know how many targets are there in a given environment which presents additional challenges to existing strategies. 
        
        Our primary contribution is an integrated solution for the distributed online multi-agent multi-target search problem, where agents only have access to limited relative state information based on local sensing and communication between neighbors in an r-disk. Our solution combines a distributed virtual pheromone-based storage and waypoint selection algorithm for coverage control, with a distributed greedy target selection algorithm to ensure the team cooperatively finds all the targets in the domain. We had presented an initial concept of the  algorithm and preliminary simulation results in our previous work~\cite{Mathew2022}, and in this paper, we further extend it with refinements to the algorithm, study of the performance of the algorithm in multiple scenarios, and the proof-of-concept validation of the proposed solution in real robots. The paper also details the real-world sensing suite and the data processing pipeline for the implementation of the solution on a robotic multi-agent system consisting of LTA agents that attempt to search and track multicolored balls in a 3D environment.

    %%%%%%%%%%%%%%%%%%%%%%%%%%%%%%%%%%%%%%%%%%%%%%%%%%%%%%%%%%%%%%%%%%%%%%%%%%%%%%%%%%%%%%%%%%%%%%%%%%%%%%%%%%%%%%%%%%%%
    % Problem Statement
    %%%%%%%%%%%%%%%%%%%%%%%%%%%%%%%%%%%%%%%%%%%%%%%%%%%%%%%%%%%%%%%%%%%%%%%%%%%%%%%%%%%%%%%%%%%%%%%%%%%%%%%%%%%%%%%%%%%%

    \section{Problem Formulation}
    \label{sec:ProblemStatement}

        Consider an environment~$\Omega \subset \real^{d_p}$ with~$d_p \in \{2, 3\}$. There are $\robotsn$ robots~$\robotpos \in SE(d_p), \; i \in \{1, ..., N_s\} = \robots$ and $\targetsn$ targets in the environment~$x_k(t) \in \real^{d_h}, \; k \in \{1, ... , N_h\} = \targets$. The robots have the ability to identify a target of interest but have no prior knowledge of where the targets are or how many targets are in the environment. We assume that $\robotsn \geq \targetsn$. The objective is to find and track these targets. 

        We consider general discrete-time nonlinear dynamics     
            \begin{align}
            \label{eqn:AgentDynamics_SingleInt}
            \begin{bmatrix}
                \robotpos(t + 1), 
                \robotvel(t + 1)
            \end{bmatrix}^T = \robotdynams (\robotpos(t), \robotvel(t), \robotinput(t)),
        \end{align}
        for a given time-step~$\timestep > 0$. $\robotvel$ denotes the velocities of the agent and $\robotinput$ is the control input for the robot. The robots do not have access to their global positions at any time. 
                
               The targets can move according to any motion model that is unknown to the agents. In this paper, for simplicity, we choose brownian motion defined as 
        \begin{align}
            \label{eqn:TargetDynamics_Linear}
            \begin{aligned}
                \targetstate(t + 1) &= \targetstate(t) + \targetprcsnoise(t), \\ \targetprcsnoise(t) &\thicksim \mathcal{N}(0, \targetprcscov(t)).
            \end{aligned}
        \end{align}
	        
        For any target motion model, the following maximum speed assumption holds:

        \begin{assumption}[Target Max Speed]
            \label{asn:AgentTarget_SpeedRelation}
            {\rm The maximum possible speed of the target is bounded by $\vmax^x$;  $\TwoNorm{x_k(t+1)-x_k(t)} \leq \vmax^x \Delta t$. Hence, we have an upper bound $\targetprcscovmax$ on the covariance of the process noise which is known.  
            }
        \end{assumption}
        
        We will now describe the most general minimum capabilities required in terms of actuation, communication, and sensing that can be applied to any particular dynamics. Later in Section~\ref{sec:toRealRobots} we show exactly how we realize all these capabilities on a real multi-robot system to validate our results. 
        
        \subsection{Actuation}

            The minimum actuation capabilities needed for our robots is a way to generate a control input~$u_i$ to move the robot faster than the targets, and (at least loosely) follow general waypoints in the domain~$\Omega$. This is formalized as: 

            \begin{assumption}[Agent Max Speed]
                \label{asn:MaxVelocity}
                {\rm The maximum possible speed of a given agent is bounded by $\vmax^p$;  $\TwoNorm{\robotpos(t+1) - \robotpos(t)} \leq \vmax^p \Delta t$. Additionally, the limit on the speed of the agent is more than that of the target, $\vmax^p > \vmax^x.$}
            \end{assumption}

            \begin{assumption}[Waypoint Control]
                \label{asn:WaypointValidity}
                {\rm For any waypoint defined~$\robotwayptabs \in \Omega$, there exists a realizable control~$\robotinput(t)$ for agent~$i$ that can move the robot closer to the waypoint in one time-step without leaving the domain; i.e.,~$\lVert \robotpos(t + 1) - \robotwayptabs \rVert < \lVert \robotpos(t) - \robotwayptabs \rVert$ and $[(\robotpos(t+1)), (\robotpos(t))] \subset \Omega.$ }
            \end{assumption}

        \subsection{Communication}
            Each agent~$i$ can broadcast information to neighboring agents within a distance~$\commrange > 0$. This is a commonly found communication model~\cite{Viseras2020,More2016}. We assume agents are able to send any locally available variables and specifics of the actual data being sent will be discussed further in section \ref{sec:DistributedActivePerception}. 
       
            \begin{assumption}[$r-$disk Communication]
                \label{asn:communication}
                {\rm Each agent~$i \in \robots$ can receive information from neighbors~$j \in \robots$ whose~$\TwoNorm{p_i - p_j} \leq r_c$ for a known communication radius~$r_c~>~0$.}
            \end{assumption}

        \subsection{Sensing and Perception}\label{se:sense}

            Lastly, each robot needs some capability of measuring/estimating the state of the targets of interest, the relative positions of other robots, and its own motion. We partition this into three components: target sensing, other agent sensing, and self-sensing; all in agent local frames. 
            
		\subsubsection{Target Detection System}\label{se:detection}

        Agent $i$ can sense targets $k$ in its Field of View (FOV) which is a subset of the environment. This region is defined as $\operatorname{FOV}(p_i(t)): SE(d_p) \rightrightarrows \Omega$. To simplify notation, the $\operatorname{FOV}(p_i(t))$ is written as $\robotfov$ in the paper. We therefore define the relative state of the target $k$ with respect to agent $i$ as $x_{i, k}$. 
             
            The output that agent~$i$ receives about target~$k$ is given by
            \begin{align}
                \label{eqn:HiddenStateMeasurement_Sensing}
                \begin{aligned}    
                    &\targetsensraw (t) = \begin{cases}
                        \targetdetection(\robotpos(t), \targetstate(t)) + \targetmeasnoise(t), \; \text{if} \; \targetstate \in \robotfov(t), \\
                        \emptyset, \quad \text{otherwise}, 
                    \end{cases} \\
                    &\targetmeasnoise(t) \thicksim \mathcal {N}(0, \targetsenscovmap(\targetstaterel)), 
                \end{aligned}    
            \end{align}
            where~$\targetdetection: \real^{d_p} \times \real^{d_h} \to \real^{d_s}$ maps the current state of robot~$i$ and target state~$k$ to an estimate of the target via a detection system for each target~$k \in \targets$. In the case that state~$p_i(t)$ does not allow robot~$i$ to sense anything about target~$k$, then~$\targetsensraw = \emptyset$. As a simplifying assumption, all targets are uniquely identifiable by the detector.

			We assume additive noise~$\targetmeasnoise(t)$, where~$\targetsenscovmap(\targetstaterel), \targetsenscovmap: \real^{d_p} \to \real^{d_h \times d_h}$ is a covariance map of sensor noise intensities at various regions of the FOV. This covariance map $\targetsenscovmap$ provides the covariance of the measurement of target~$\targetsenscov$. We show how this can be determined by experimental methods of sensing the target at various positions in the FOV and extrapolating the measured noise over the entire region in appendix~\ref{se:detectiondetails}. Using the raw estimate of the target~$\targetsensraw$ and the associated sensor noise covariance from the covariance map $R_i^s$, we can maintain an estimate of the target~$\targetsenspos$. Details of maintaining the target estimate are mentioned in section~\ref{se:selection}.
        
        \begin{assumption}[Sensing-Comms. Relation]
            \label{asn:FOVComm_Relation}
            {\rm For any agent, the sensing FOV region is within the communication ball, $\robotfov \subset \bar{B}(\robotpos(t), \commrange)$. }%In practice, most hardware setups satisfy this relation.} 
        \end{assumption}
 
        \subsubsection{Relative Position Sensing}
        \label{se:relativeposition}

        The global positions~$\robotpos(t)$ are not available to the agents; to compensate for this, the agents must have two other sensing capabilities. First, an agent can sense the relative position of any neighboring agent within the communication range $\neighpos = p_j - p_i$ along with an estimate of the associated uncertainty~$\neighposcov$ using a covariance map~$\neighposcovmap$. This relative position sensing must happen at least as frequent as the data transmission rates; Ideally, with each transmission, the relative position can be sensed.
        
        \subsubsection{Displacement Sensing}
        \label{se:displacement}
        
        Finally, the agents can sense their own latest relative displacement $\robotposlast(t) := \robotpos(t) - \robotpos(t-1)$ between each processing cycle. The agent can also know the associated covariance $\robotposlastcov$ from a covariance map $\robotposlastcovmap$.

        Based on the locally maintained target estimates~$\targetsenspos$ and information received from other agents~$j \neq i$ of their local estimates maintained similarly, each agent~$i$ must construct a combined estimate~$\targetsenscomb(t)$ for all targets~$k \in \targets$ it has information about. We show the exact formulation of the combined estimate later in section \ref{sec:DistributedActivePerception}. The overall goal of the multi-robot system is then to cooperatively maintain estimates about all the different targets in the domain. This is formalized in Problem~\ref{pr:activeperception}.

        \begin{problem}[Distributed Active Perception]\label{pr:activeperception}
            Given any initial configuration of mobile sensing robots~$\{p_1(0), ..., p_{N_s}(0)\} \in \Omega^{N_s},$ with dynamics~\eqref{eqn:AgentDynamics_SingleInt} and target dynamics~\eqref{eqn:TargetDynamics_Linear} satisfying Assumptions \ref{asn:AgentTarget_SpeedRelation}, \ref{asn:MaxVelocity}, \ref{asn:WaypointValidity}, \ref{asn:communication} and~\ref{asn:FOVComm_Relation}, find a distributed control strategy~$\robotinput(t)$ and estimate~$\targetsenscomb(t)$ for each~$i \in \robots$ such that the multi-robot team cooperatively reduces
            \begin{align}\label{eq:mainobjective}
                \costtracking := \frac{1}{\robotsn}\sum_{k \in \targets} \min_{i \in \robots} E[ \TwoNorm{\targetstaterel(t)-\targetsenscomb(t)} ]. 
            \end{align}
        \end{problem}

        \begin{figure*}[ht!]
            \centering
            \includegraphics[width=0.85\linewidth]{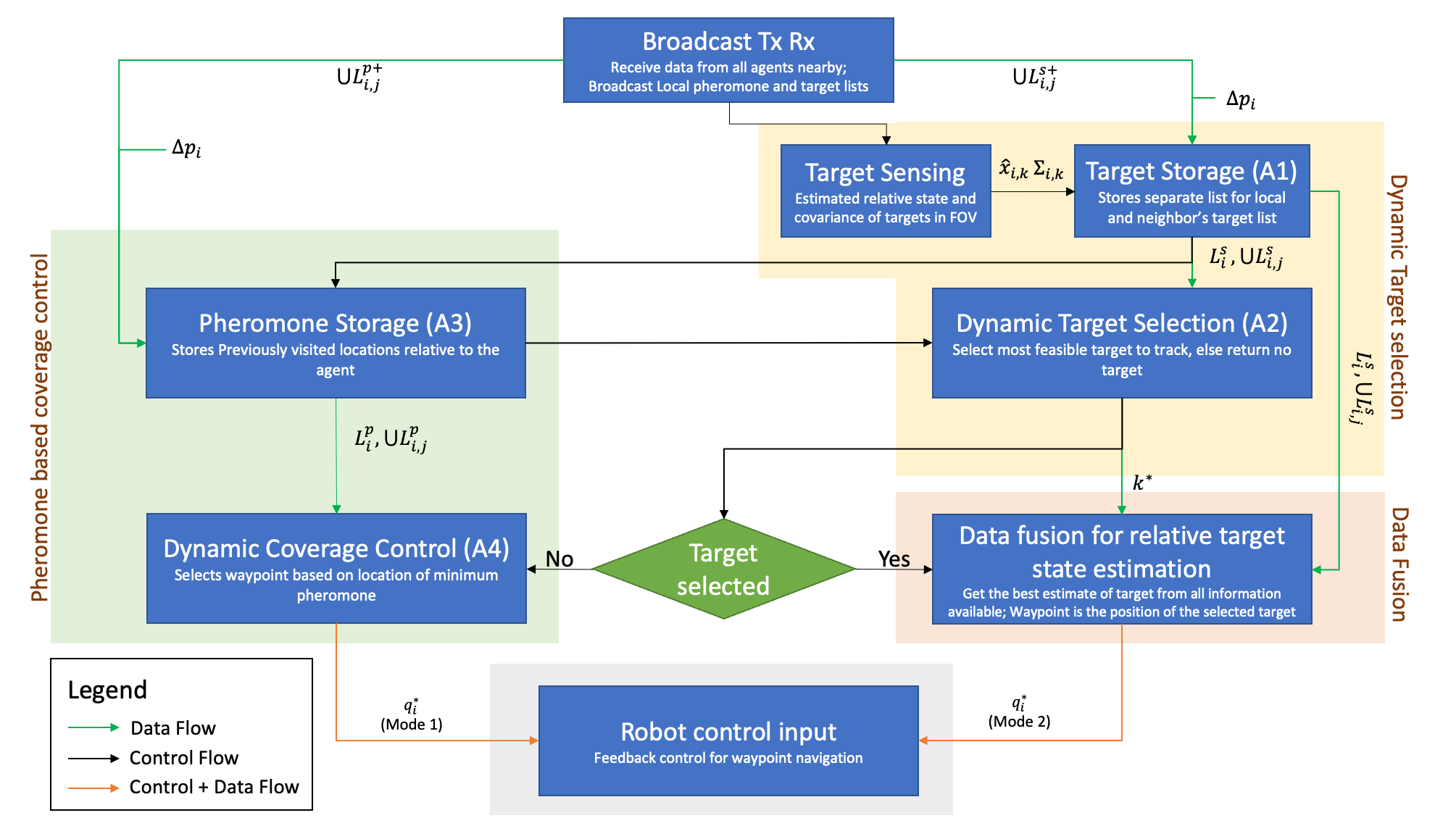}
            \caption{Flow-chart of one iteration of solution running in agent.}
            \label{fig:solution_flowcahrt}
        \end{figure*}

    %%%%%%%%%%%%%%%%%%%%%%%%%%%%%%%%%%%%%%%%%%%%%%%%%%%%%%%%%%%%%%%%%%%%%%%%%%%%%%%%%%%%%%%%%%%%%%%%%%%%%%%%%%%%%%%%%%%%
    % Distributed Active Perception
    %%%%%%%%%%%%%%%%%%%%%%%%%%%%%%%%%%%%%%%%%%%%%%%%%%%%%%%%%%%%%%%%%%%%%%%%%%%%%%%%%%%%%%%%%%%%%%%%%%%%%%%%%%%%%%%%%%%%
    \section{Distributed Active Perception}
    \label{sec:DistributedActivePerception}

    To solve the cooperative multi-robot active perception problem, we separate the problem into two more fundamental problems with known solutions, i.e., searching the environment and tracking targets. We further fuse these 2 solutions on a distributed multi-robot system
    
    First, let us imagine that there are no targets in the environment at all (but the robots do not know this). What should the system do in this case? It is easy now to connect this to a coverage control problem where we might implement a Lloyd's descent algorithm~\cite{Cortes2004}, with a few modifications for our specific case. Since there is no information on targets, we assume that the targets are equally likely to be anywhere in the environment. In-fact there are multiple search algorithms that can be plugged into this sub-problem, like Anti-flocking search~\cite{Ganganath2016}, or even simple Random-walk and Levy walk. Our virtual pheromone based solution is a modification to the Anti-flocking algorithm such that it can be deployed in GPS denied environment where agents have limited communication ranges. 

    Second, let us now imagine that the locations of all targets are generally known, but our robots now need to be properly matched up with an appropriate target in order to maintain some system-level estimates of the targets. This is very similar to a generic assignment problem~\cite{Dirk1994, Ozbakir2010}. Here however, all agents work with incomplete and often asymmetric information which brings in the challenge this sub-problem. Our solution is a ``distributed greedy'' assignment algorithm. Later in simulations, we show comparable performance against the popular Auction algorithm~\cite{bertsekas1979} for this scenario.

    In the rest of this section, we present our end-to-end pheromone-based searching and tracking algorithm. We first show how to cooperatively maintain estimates about targets using Algorithm~\ref{alg:TargetStorage} explained in Section~\ref{se:selection}. Based on these estimates the agents negotiate which targets to track as explained in Algorithm~\ref{alg:TargetChoosing}. Then, we show how agents should search the space when no good target is provided by Algorithm~\ref{alg:Pheromone_DataStorage} and Algorithm~\ref{alg:Pheromone_Combine} using a virtual Pheromone-based approach in Section~\ref{se:virtualPheromone}. Finally, in Section~\ref{se:estimation} we provide the fully synthesized distributed active perception Algorithm~\ref{alg:AgentControl_Full}. Figure \ref{fig:solution_flowcahrt} shows an overview of the solution.

    \subsection{Dynamic Target Tracking and Selection}\label{se:selection}

        Each round begins with each agent~$i$ processing any pending information communicated by neighbors and getting the relative state estimate of the targets $\hat{x}_{i, k}^{\text{raw}}$ detected in the FOV~\eqref{eqn:HiddenStateMeasurement_Sensing}. The locally sensed targets are stored as a local target list and the target data received from neighbors is stored as neighboring target list. Using the measure of uncertainty 
        \begin{align}
            \label{eqn:TargetMeasurement_FOV_DiffEntrpy}
                    &\targetentropylocal = \det \Sigma_{i, k} (t),
        \end{align}
        from the local target detection and the received neighbor target lists, the agents determine the target that needs to be tracked using a pseudo-target negotiation. 
        
        \subsubsection{Target Storage}
            When an agent detects a target in its FOV, it computes the estimated state $\targetsenspos(t)$ and the covariance of estimation $\targetsenscov$ as seen in~\eqref{eqn:TargetMeasurement_FOV}. After a target~$k$ is initially observed, maintaining an estimate of~$\targetsenspos$ will depend on whether the target remains or leaves the FOV. In each step we update the relative position of the target with the motion of the agent $\robotposlast$ in the previous time-step. Since the only information we have on the target is the maximum bound on the covariance growth $\targetprcscovmax$, we use this bound during the covariance propagation. With any new observation of target, its estimate is propagated forward in time using the error ellipses method~\cite{Blacciiman1989}, and the uncertainties are propagated via covariance propagation~\cite{Grewal2001} using    
            \begin{subequations}
                \allowdisplaybreaks
                \label{eqn:TargetMeasurement_FOV}
                \begin{align}
                    \label{eqn:TargetMeasurement_FOV_Sigma}
                    &\Sigma_{i, k}^-(t) = \Sigma_{i, k} (t - 1) + \bar{Q}_k, \\
                    \label{eqn:TargetMeasurement_FOV_Sigma_Raw}
                    &\Sigma_{i, k}^+(t) = R_i^s(\hat{x}_{i, k}^{\text{raw}} (t)),  \\
                    \label{eqn:TargetMeasurement_FOV_Covariance}
                    &\Sigma_{i, k} (t) =  \begin{cases}
                        \left(\Sigma_{i, k}^+(t)^{-1} + \Sigma_{i, k}^-(t)^{-1}\right)^{-1}, \\ \quad \quad \text{if} \; x_k(t) \in \text{FOV}_i, \\
                        \Sigma_{i, k}^-(t), \; \text{otherwise},
                    \end{cases}\\
                    \label{eqn:TargetMeasurement_FOV_Estimate}
                    &\hat{x}_{i, k}(t) = \begin{cases}
                        \Sigma_{i, k} (t) \left(\Sigma_{i, k}^+(t)^{-1} \hat{x}_{i, k}^{\text{raw}}(t) + \right.\\
                        \quad \quad \left.\Sigma_{i, k}^-(t)^{-1} \left(\hat{x}_{i, k}(t - 1) + \Delta p_i(t)\right) \right), \\ 
                        \quad \quad \text{if} \; x_k(t) \in \text{FOV}_i, \\
                        \hat{x}_{i, k}(t - 1) + \Delta p_i(t), \; \text{otherwise}.
                    \end{cases}
                \end{align}
            \end{subequations}

            The sensed targets get stored in a local target list $\senslistlocal$. In addition to targets directly sensed by agent~$i$ in each round it receives the local target list from neighbors~$j$ defined as $\senslistrec$ to update its neighbor target list~$\senslistneig$. This consists of the neighbor's sensed targets $\targetsensneighpos$, and the associated covariance $\targetsensneighcov$. When there is no data received from the neighbors, the agent simply grows the uncertainty of the targets. 
            
            We also assume that the process of transmission of the target lists allow the sensing of the relative position to neighbor~$\neighposrec$. Any new update from the neighbors will update the position of the neighbor using the previously mentioned error ellipses method as 
            \begin{subequations}
                \label{eqn:Agent2Agent_Propogation}
                \begin{align}
                    &\neighposcov^+(t) = R_i^p(\neighposrec(t)),  \\
                    \label{eqn:Agent2Agent_Covariance}
                    &\neighposcov (t) = \left(\neighposcov(t - 1)^{-1} + \neighposcov^+(t)^{-1}\right)^{-1} \\ \nonumber
                    \label{eqn:Agent2Agent_Estimate}
                    &\neighpos (t)  = \neighposcov (t) \left(\neighposcov^+(t)^{-1} \neighposrec(t) + \right.\\
                    &\quad \quad \quad \quad \left.\neighposcov(t - 1)^{-1} \neighpos (t - 1) \right).
                \end{align}
            \end{subequations}

            The target storage system takes care of covariance propagation and location updates of the targets in each sensing list based on new information received from the sensed targets and target data received from neighbors. We formalize all the functionalities of target storage system in Algorithm~\ref{alg:TargetStorage}, where, in steps~\algostep{2-5}, we show the updates to the local target list, and in steps~\algostep{6-14}, the updates to the target lists of neighbors. Additionally, in steps ~\algostep{5, 14}, it also takes care of removing elements with large uncertainties over a threshold $\targetcovmax$, which denotes very low confidence in the last known state of the target.
            
            \begin{algorithm}[t]
                \caption{Detection and Tracking: Target Storage}
                \label{alg:TargetStorage}
                \hspace*{\algorithmicindent} \textbf{Input} Targets sensed in FOV, Received Target List $\senslistrec$ \\
                \hspace*{\algorithmicindent} \textbf{Output} Target Lists $\senslistlocal, \cup \senslistneig$ and relative positions of neighbors $\cup \neighpos,\cup \neighposcov$
                \begin{algorithmic}[1]
                    \State Get last relative position $\robotposlast := \robotpos(t-1) - \robotpos(t)$ and the associated covariance $\robotposlastcov$.
                    \For{$(\targetsenspos, \targetsenscov) \in \senslistlocal$ and detected $\targetsensraw$}
                        \State Propagate target relative position $\targetsenspos$ and the target covariance $\targetsenscov$ using \eqref{eqn:TargetMeasurement_FOV} and update local target list $\senslistlocal$ with those values.
                    \EndFor
                    \State Add detected target not in $\senslistlocal$ with the current relative position and instantaneous covariance shown in \eqref{eqn:TargetMeasurement_FOV_Sigma_Raw}. 
                    \State Delete all $l_{i, k}^s \in \senslistlocal$ with large uncertainty above a threshold $\det\Sigma_{i, k} > \bar{\Sigma}_i$.

                    \For {$j \neq i$ that broadcast data $\senslistrec$ was received from}
                        \State Set $\senslistneig \gets$ latest agent target list from agent $j$.
                        \State Update $\neighpos$ and compute corresponding $\neighposcov$ using the error ellipsis method \eqref{eqn:Agent2Agent_Propogation}.
                    \EndFor

                    \For {$j \neq i$ that no info was received from}
                        \If{ previous data exists $\senslistneig \neq \emptyset$ }
                            \State Update neighboring agent position $\neighpos(t) \gets \neighpos(t - 1) + \robotposlast(t)$ with local agent displacement.
                            \State Propagate neighboring agent position covariance $\neighposcov \gets \neighposcov + \robotposlastcov$.   
                            \State Propagate target covariance $\targetsensneighcov (t) \gets \targetsensneighcov (t - 1) + \bar{Q}_k$ for all targets in $\senslistneig$.
                        \EndIf
                    \EndFor
                    \State Delete all $l_{i, j, k}^s \in \senslistneig$ with $\det \Sigma_{i, j, k} > \bar{\Sigma}_i$.

                    \Return $\senslistlocal, \senslistneig, \cup \neighpos, \cup \neighposcov$
                \end{algorithmic}
            \end{algorithm}
            
            \begin{algorithm}[t]
                \caption{Detection and Tracking: Target Selection}
                \label{alg:TargetChoosing}
                \hspace*{\algorithmicindent} \textbf{Input} Local Target List $L_i^s$, Neighbor target lists $\cup L_{i, j}^s$, relative position between neighbors $\cup \hat{p}_{i, j}, \cup \Sigma_{i, j}$ \\
                \hspace*{\algorithmicindent} \textbf{Output} Agent target selection $k_i^* > 0$ or defer to exploration mode $k_i^* = 0$
                \begin{algorithmic}[1]
                    \State Initialize~$\tilde{K}_i \subset \targets$ set of candidate targets with estimates~$\hat{x}_{i,k} \neq \emptyset$ or~$\hat{x}_{i,j,k} \neq \emptyset$ 
                    \If{$\tilde{K}_i = \emptyset$}
                        \State $k_i^* \gets 0$.
                    \Else
                        \For {$m \in \robots$} 
                            \For {$l_m^s = (\hat{x}_{m, k}, \Sigma_{m, k}) \in L_m^s$ sorted by $\Sigma_{m, k}$} 
                                \State Compute $\targetentropylocalnv{m}$ using \eqref{eqn:TargetMeasurement_FOV}. 
                                \State Compute all other $\targetentropylocalnv{\bar{m}}$ using the target list corresponding to not $m$ using \eqref{eqn:TargetMeasurement_FOV}. 
                                % \State Using Hungarian algorithm, perform pseudo assignment of targets to agents using the computed entropies.  
                                \If{$\targetentropylocalnv{m}$ is the lowest}
                                    \State Assign target $k$ to agent $m$ and end assignment for $m$.
                                \EndIf
                            \EndFor 
                        \EndFor
                        \If{Valid assignment for agent $i$}
                            \State $k_i^* \gets$ target assigned to $i$.
                        \Else
                            \State $\bar{K}_i \gets$ targets without valid assignment in \algostep{5-9}.
                            \If{$\bar{K}_i = \emptyset$}
                                \State $k_i^* \gets 0$.
                            \Else
                                \State Compute entropy $\mathbf{h}(x_{i, k}|\hat{x}^{\text{raw}}_{j, k}(t))$ for all targets in $\bar{K}$ using \eqref{eqn:InformationEntropy_Neighbour}.
                                \State $k_i^* \gets $ target $k \in \bar{K}$ with least $\mathbf{h}(x_{i, k}|\hat{x}^{\text{raw}}_{j, k}(t))$.
                            \EndIf
                        \EndIf
                    \EndIf
                    \Return $k_i^*$
                \end{algorithmic}
            \end{algorithm}

        \subsubsection{Target Selection} 
        \label{se:assignment}
\begin{figure}[h]
                \centering
                \includegraphics[width=\linewidth]{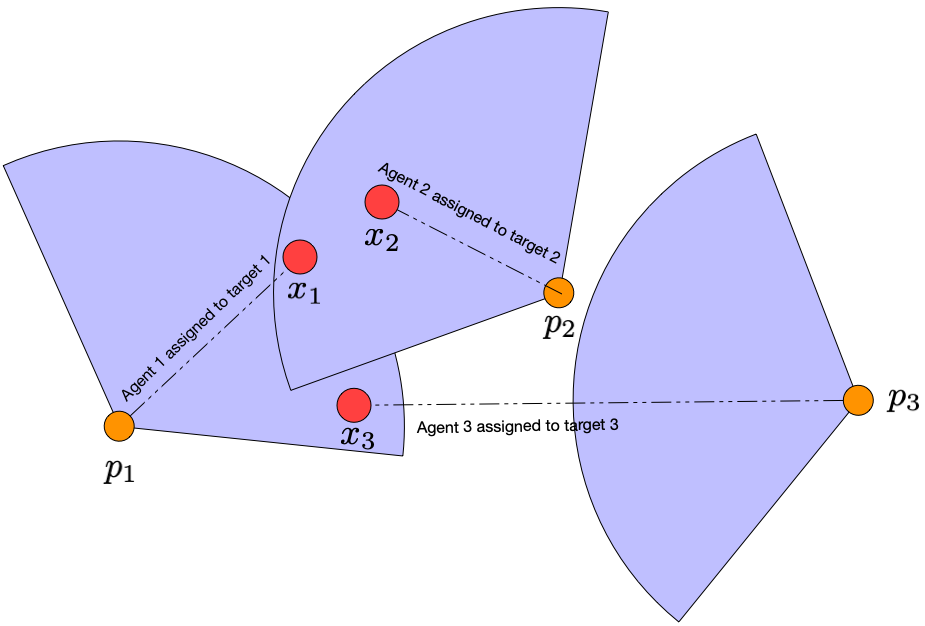}
                \caption{Target selection strategy. In this illustration, we show how the targets will be assigned when 2 or more targets are in the FOV (Agents 1 and 2) and how a target can be assigned in a neighboring agent's FOV (Agent 3). }
                \label{fig:TargetSelection}
            \end{figure}

            With the targets segregated in local target list and the neighbor target list, each agent can make a pseudo-decision on which target to track based on the target available to that agent at the time. We will term this target assignment strategy a ``distributed-greedy'' target assignment strategy. This is divided into two phases. In the first phase of target assignment, an agent can only choose a target in its own FOV. The choice of target is done such that the agent has the least local uncertainty of that target in comparison to all other agents. This is done to prioritize not losing targets already in FOV since it constitutes an increase in the overall uncertainty of the target. We show a hypothetical assignment in Figure~\ref{fig:TargetSelection}. In this round, if a target~$\targetchosen$ was assigned to agent~$i$, then the assignment ends with that chosen target.

            If however a target was not assigned in this first round assignment, we choose an unassigned target in a neighbor's FOV with the least uncertainty as shown in Figure~\ref{fig:TargetSelection}. Since all agents work with relative positions referenced to its body frame, we need to translate the targets in the neighbor's target list from the neighbor's body frame $\targetsensneighpos$ to the local agent's body frame $\targetsensneighlocalpos$. In the process, the estimate incurs an additional source of uncertainty coming from the noise in the relative positions between agents themselves as 
                \begin{subequations}
                    \label{eqn:InformationEntropy_Neighbour}
                    \begin{align}
                        \hat{x}_{i, k}^j(t) &= \hat{x}_{i, j, k}(t) + \hat{p}_{i, j}(t), \\
                        \Sigma_{i, k}^j (t) &=  \Sigma_{i, j, k}(t) + \Sigma_{i, j}(t)\\
                    \mathbf{h}(x_{i, k}|\hat{x}^{\text{raw}}_{j, k}(0:t)) &= \det \Sigma_{i, k}^j (t).
                    \end{align}
                \end{subequations}
            Since the target sensing and neighboring agent relative position  sensing uses independent data sources, we assume no correlation between them. This relative position and the associated uncertainty is used in the second round assignment, where the agent chooses the target with the least uncertainty~$\targetentropylocalnv{j}$.

            We formalize the distributed target tracking solution in Algorithm~\ref{alg:TargetChoosing}. With the local and the neighbor target lists, we start the target selection process, where, in the first round of assignment shown in steps~\algostep{5-12}, the local uncertainty is used to derive the target assignment. Subsequently, if agent $i$ did not get an assignment previously, steps~\algostep{14-19} shows the assignment to a target in neighbor's FOV.

        \subsection{Virtual Pheromone Storage and Mapping}\label{se:virtualPheromone}
            For each agent~$i$ for which no target has been assigned~$k_i^* = 0$, it should choose its input~$u_i(t)$ to maximize exploration of the environment. Here we present a biologically inspired pheromone-based algorithm, that determines how an agent searches for undiscovered targets, taking inspiration from some existing pheromone based systems~\cite{Clavio2011, Oliveira2014, Sun2019, Arvin2018} modified to be fully distributed. Each agent stores the relative position of previously visited locations. Since this must be done in the relative frame, in each round, agent~$i$ generates a new virtual pheromone $q_{i, s}^p$ at the point it had last visited relative to its current position~$\Delta p_i(t) := p_i(t-1) - p_i(t)$, with an associated maximum weight $\pherweightinit$. Most pheromone systems \cite{Sun2019, Arvin2018} uses an exponential decay for pheromones. Hence, the time-varying weight~$w_{i, s}^p(t)$ for each pheromone deposit decays at a rate of~$\pherweightdecay$ per time-step as $$w_{i, s}^p(t + 1) = w_{i, s}^p(t)\left(1-\pherweightdecay\right)$$ with a lower limit of $\pherweightthresh$. These virtual pheromones will then be shared among the agents in a distributed fashion to allow the agents to cooperatively search for targets in places the team collectively has not yet visited. Intuitively, the decaying weights track areas of the domain that were most recently searched and so don't need to be revisited again for a while. Combining the pheromones created locally, and the ones obtained from neighbors, an agent can create a pheromone map to then figure out the least visited location and move there. To ease the memory burden, we store only the relative position it was at in the previous time-step and weight~$w_i^p > \pherweightthresh$.

            In short, in each communication round the agents generate and/or receive new pheromones while propagating the existing pheromones forward in time. The algorithm returns both a pheromone list for its own pheromones $L_i^p$ and the collective pheromone list of neighboring agents $\cup L_{i, j}^p$. This is formalized in Algorithm~\ref{alg:Pheromone_DataStorage} where steps~\algostep{6} generates the new pheromone, and steps~\algostep{2-5,7-12} propagate these decaying pheromones forward in time both locally to agent~$i$ and its neighbors, respectively.

            \begin{algorithm}[t]
                \caption{Virtual Pheromone: Pheromone Storage}
                \label{alg:Pheromone_DataStorage}
                \hspace*{\algorithmicindent} \textbf{Input} Received Pheromone List $L_{i, j}^{p+}$ \\
                \hspace*{\algorithmicindent} \textbf{Output} Pheromone Lists $L_i^p, \cup L_{i, j}^p$
                \begin{algorithmic}[1]
                    \State Get last relative position $\robotposlast (t)$ and the associated covariance $\robotposlastcov (t)$.
                    \For{$l_{i,s}^p = (q_{i,s}^p, \phercov, w_{i,s}^p) \in L_i^p$}
                        \State Propagate relative position $q_{i,s}^p \gets q_{i,s}^p + \Delta p_i(t)$. 
                        \State Propagate covariance $\phercov \gets \phercov + \robotposlastcov (t)$.
                        \State Pheromone strength decays $w_{i, s}^p \gets \max \{ w_{i,s}^p(1- \pherweightdecay), 
                        \pherweightthresh \} $; delete pheromone if weight is $\pherweightthresh$.
%                            \If {$w_i^p \leq 0$}
%                                \State Delete $l_i^p$ from $L_i^p$
%                            \EndIf
                    \EndFor
                    \State Generate a new element for pheromone list and add $\left(\Delta p_i(t), \robotposlastcov (t), w_i^{\text{init}}\right)$ to~$L_i^p$.
                    %\State Add $l_i^{p+}$ to $L_i^p$
                    \For {$j \in \robots \setminus \{i\}$}
                        \If{Pheromone List from Agent $j$ received}
                            \State Overwrite $L_{i, j}^p$ with received list $L_{i, j}^{p+}$. 
                        \Else
                            \For{$l_{i, j}^p = (q_{i, j, s}^p, \pherneighcov w_{i, j, s}^p) \in L_i^p$}
                                \State $w_{i, j, s}^p \gets \max \{w_{i, j, s}^p(1 - \pherweightdecay), \pherweightthresh\}$; Delete pheromone if weight is $\pherweightthresh$.
%                                    \If {$w_{i, j}^p \leq 0$}
%                                        \State Delete $l_{i, j}^p$ from $L_{i, j}^p$
%                                    \EndIf
                            \EndFor
                        \EndIf
                    \EndFor
                    \Return $L_i^p, \cup L_{i, j}^p$
                \end{algorithmic}
            \end{algorithm}
            
            Given this historical information of its own pheromones along with the received pheromones of neighbors, each agent can construct the combined pheromone list $L_i^c$. This list has pheromone regions $Q_{i, s}^p$ derived from the position of the pheromone in memory defined by 
            \begin{align}
                \label{eqn:PheromoneFOV}
                Q_{i, s}^p(q) &:= \robotfovshrink (q, \operatorname{FOV}_i(q_{i, s}^p), \phercov).
            \end{align}
            Here $\robotfovshrink: \Omega \to \realnonnegative$ diffuses the FOV defining the pheromone area based on the uncertainty in the pheromone position by performing a convolution of the FOV region with a Gaussian kernel corresponding to the covariance $\phercov$. Using this diffusion, the areas with lower pheromone weight are guaranteed to be unexplored.
            
\begin{figure}[ht]
                \centering
                \includegraphics[width=0.9\linewidth]{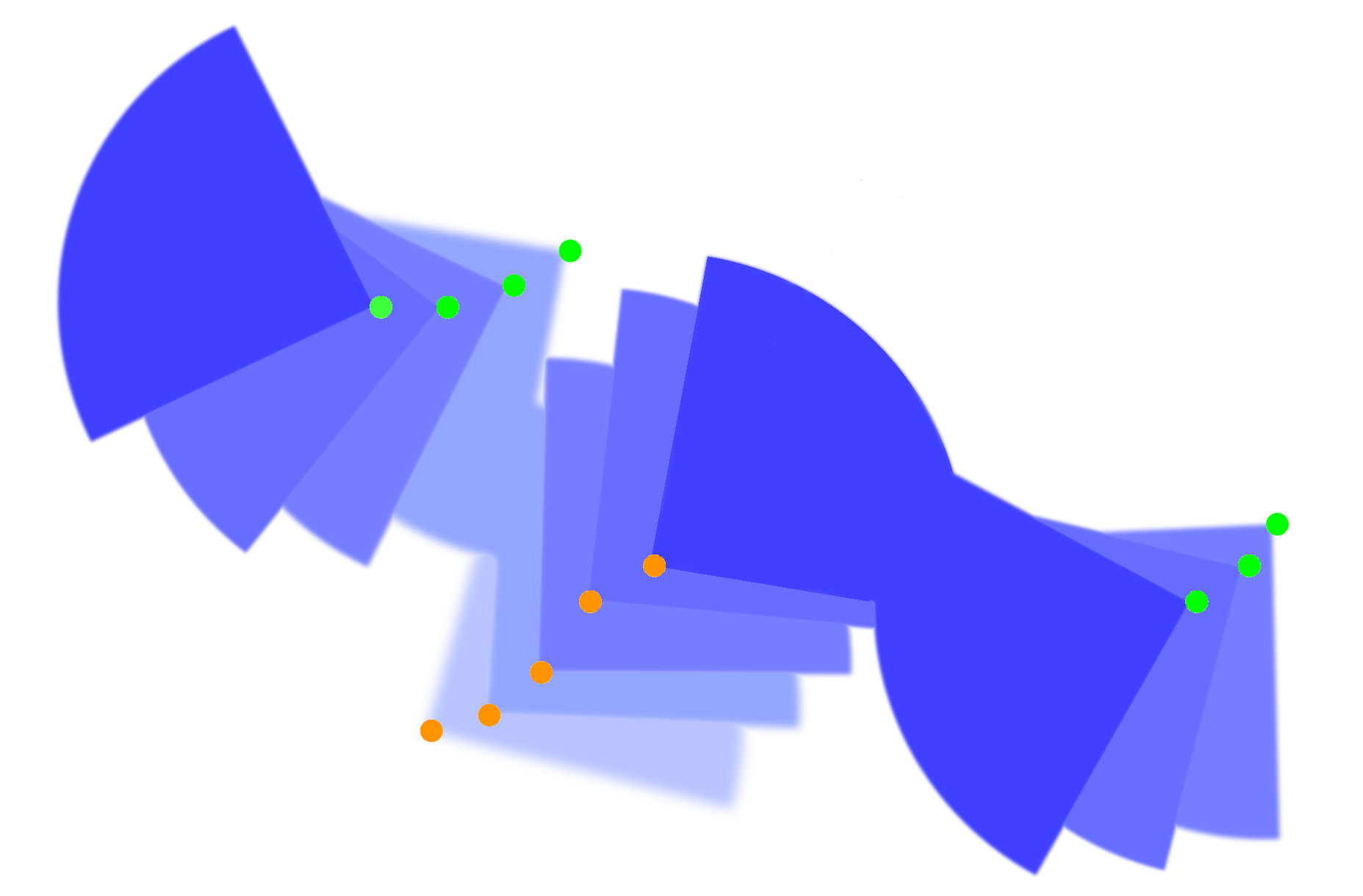}
                \caption{Pheromone Map. Orange shows the pheromones in the local pheromone list. Green points show the locations of pheromone in the neighbor pheromone list. The different shades of blue shows the different strengths of pheromone.}
                \label{fig:Target_Mode1}
            \end{figure}

            Finally, the pheromone map $m_i^p(q): \Omega \to \realnonnegative$ for an agent defines the regions around the agent that it or its neighboring agents have visited in the past with the strength of the pheromone in the region defining how recent this visit was. It is defined as 
            \begin{align}
                \label{eqn:PheromoneMap}
                m_i^p(q) &:= \max_{L_i^c} \{Q_{i, s}^p (q)\}.
            \end{align}
            To generate the pheromone map all the pheromone regions in the combined pheromone list are superimposed on a single map. Further, after the superimposition, the map is flattened by taking the maximum pheromone weight out of all the superimpositions as shown in \eqref{eqn:PheromoneMap}. A hypothetical pheromone map is shown in Figure~\ref{fig:Target_Mode1}. 

            \begin{algorithm}[t]
                \caption{Virtual Pheromone: Exploration Mode Waypoint Selection}
                \label{alg:Pheromone_Combine}
                \hspace*{\algorithmicindent} \textbf{Input} Local Pheromone List $L_i^p$, Pheromone list from neighbors $L_{i, j}^p$ and relative distance to neighbors $\hat{p}_{i, j}$\\
                \hspace*{\algorithmicindent} \textbf{Output} Waypoint for feedback control $q_i^*$
                \begin{algorithmic}[1]
                    \State Initialize combined list: $L_i^c \gets \emptyset$
                    \For {$l_{i, s}^p \in L_i^p$}
                        \State Compute the FOV region $Q_{i,s}^p$ using \eqref{eqn:PheromoneFOV} for pheromone $l_{i, s}^p$ and add it to $L_i^c$
                    \EndFor
                    \For {$j \in \robots \setminus \{i\}$} 
                        \For {$l_{i, j, s}^p \in L_{i, j}^p$}
                            \State Transform the relative position of pheromone to agent $i$: $q_{i, s}^{p, j} = q_{i, j, s}^p + \hat{p}_{i, j}$
                            \State Compute the FOV region $Q_{i, s}^p$ using \eqref{eqn:PheromoneFOV} for pheromone $l_{i, j, s}^p$ and add it to $L_i^c$
                        \EndFor
                    \EndFor
                    \State Compute the pheromone map $m_i^p(q)$ using \eqref{eqn:PheromoneMap}
                    \State Compute the waypoint $q_i^*$ using \eqref{eqn:Mode1_Waypoint}
                    
                     \Return $q_i^*$
                \end{algorithmic}
            \end{algorithm}

        \subsection{Target State Estimation and Control}\label{se:estimation}
            Finally, we show how to synthesize the solutions to the various sub-problems above to ultimately construct an estimate~$\hat{x}_{i,k^*}^c(t)$ for some chosen target~$k^*$ and how to move~$u_i(t)$. First, in the case that there is no assigned target~$k_i^* = 0$, we say agent~$i$ is in exploration mode. Otherwise, it's in exploitation mode. Both modes generate a waypoint $q_i^*$ for the robot.

            \subsubsection{Exploration Mode Waypoint Selection}    

                The main aim of the agent in this mode is to cover as much of the unexplored area as possible thereby maximizing the chance of finding potential targets as shown in Algorithm~\ref{alg:Pheromone_Combine}. Using the pheromone map $m_i^p(q, t)$, we choose a relative waypoint $q_i^*$ with the least weight within a ball around the agent as

                \begin{subequations}
                    \label{eqn:Mode1_Waypoint}
                    \allowdisplaybreaks
                    \begin{align}
                        q_i^{*-}(t) &= q_i^*(t - 1) + \Delta p_i(t), \\
                        q_i^{*+}(t) &\thicksim \mathcal{U}\{\argmin_{q \in \bar{B}(0, r_c)} m_i^p(q, t)\}, \\
                        \label{eqn:Mode1_Waypoint_recompute}
                        q_i^*(t) &= \begin{cases}
                            q_i^{*+}(t), \; \text{if} \; m_i^p(q_i^{*-}(t)) > w_{q_i^*}^p(t - 1) \\ \quad \text{or} \TwoNorm{q_i^{*-}(t)} < \bar{q}^*, \\
                            q_i^{*-}(t), \; \text{otherwise},
                        \end{cases} \\
                        w_{q_i^*}^p(t) &= m_i^p(q_i^{*}(t)).
                    \end{align}
                \end{subequations}
                Here $\mathcal{U}$ denotes a uniform distribution. The waypoint is recomputed \eqref{eqn:Mode1_Waypoint_recompute} every time the agent is near the current waypoint or the pheromone weight at the current waypoint has increased.

                The pheromone based search system intuitively directs the agents to areas of the least pheromones i.e. areas that are unexplored very recently. 
               
            \subsubsection{Exploitation Mode Waypoint Selection}    
                With the chosen target $\targetchosen$, we can get the best state estimate $\hat{x}_{i, k^*}^c(t)$ for it by fusing the target estimates from the local and the neighbor target lists using the Error ellipses method \cite{Blacciiman1989} as
                \begin{subequations}
                    \label{eqn:Target_DataFusion}
                    \begin{align}
                        &\Sigma_{i, k^*}^c (t) = \left(\Sigma_{i, k^*} (t)^{-1} + \sum_{L_{i, j}^s} \Sigma_{i, k^*}^j (t)^{-1} \right)^{-1}, \\
                        \begin{split}
                            &\hat{x}_{i, k^*}^c(t) =  \Sigma_{i, k^*}^c (t) \left(\Sigma_{i, k^*}(t)^{-1}\hat{x}_{i, k^*}(t) + \right.\\ 
                            & \quad \quad \left.\sum_{L_{i, j}^s} \left(\Sigma_{i, k^*}^j (t)\right)^{-1} 
                                \left(\hat{x}_{i, k^*}^j(t)\right) \right), 
                        \end{split} \\
                        &\mathbf{h}(x_{i, k^*}|\mathcal{I}_i(0:t)) = \det \Sigma_{i, k^*}^c (t).
                    \end{align}    
                \end{subequations}
                 
                \begin{figure}[ht]
                    \centering
                    \includegraphics[width=0.7\linewidth]{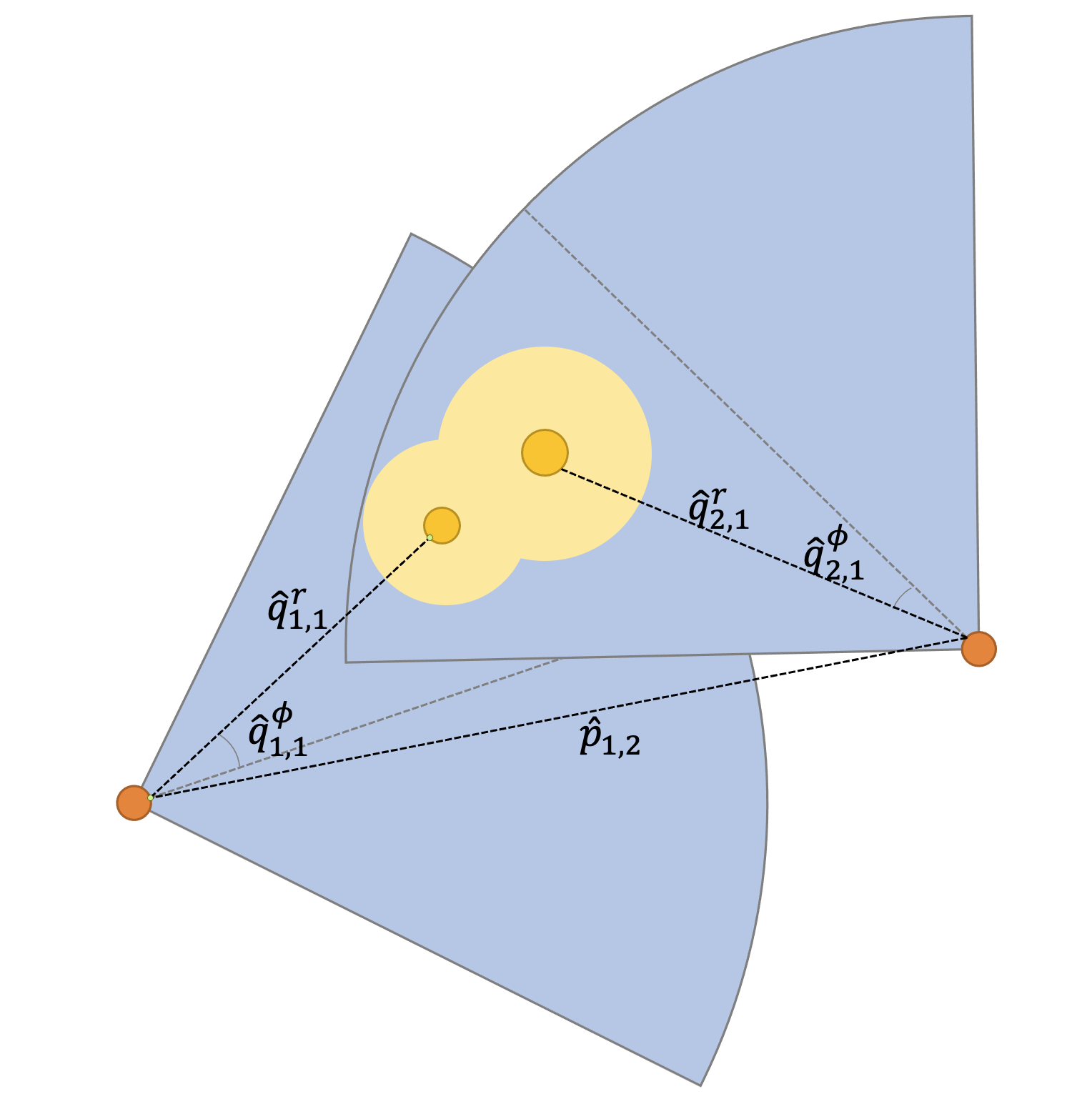}
                    \caption{Final target estimate is combined estimate of the data available in each agent. Here agent 1 combines its own estimate with that received from agent 2 using error ellipsis method}
                    \label{fig:TargetFinalEstimate}
                \end{figure}

                Using the covariance map $R_i^s(q)$, we can determine the best relative position between the agent and the target to get the best estimate. The waypoint is set such that this target eventually reaches this relative position using 
                
                \begin{align}
                    \label{eqn:Mode2_Waypoint}
                    q_i^*(t) = \hat{x}_{i, k_i^*}^c(t) - \argmin_{q \in \operatorname{FOV}_i(t)}R_i^s(q).
                \end{align}

            Once the agent determines a relative waypoint $q_i^*(t)$ to get to, the control law~$u_i(t)$ is a feedback controller to bring the relative waypoint $q_i^* \to 0$. The waypoint tracking will be very easy to integrate to this proposed solution, with choices ranging from a simple proportional controller~\cite{Panos2007} to complex adaptive control strategies~\cite{Tao2008} depending on the application.

            \begin{algorithm}[t]
                \caption{Distributed Active Perception: Agent control}
                \label{alg:AgentControl_Full}
                \begin{algorithmic}[1]
                    \State Initialize agent $i$ with position $p_i(0)$ 
                    \State Initialize empty local pheromone list $L_i^p$ 
                    \State Initialize empty neighbor pheromone lists $\cup_{j \in \robots \setminus i}L_{i, j}^p$
                    \State Initialize empty target list $L_i^s$ 
                    \State Initialize empty target list $L_{i, j}^s$ for agent $j \in \robots \setminus i$
                    \State Initialize relative distance to agent $j$: $\hat{p}_{i, j} \gets 0$ for agent $j \in \robots \setminus i$
                    \For {$t \in \integernonnegative$} 
                        \State Rx data from all available ~$j \in \robots \setminus i$
                        \State Broadcast $\mathcal{I}_i^l(t) = (L_i^p, L_i^s, \hat{p}_{i, j})$
                        \State Get Target Estimates using \eqref{eqn:TargetMeasurement_FOV}
                        \State Update Target Lists using Algorithm~\ref{alg:TargetStorage}
                        \State Update Pheromone Storage using Algorithm~\ref{alg:Pheromone_DataStorage}
                        \State Select target for agent $k_i^*$ using Algorithm~\ref{alg:TargetChoosing}
                        \If {$k_i^* = 0$}
                            \State Exploration Mode control: Compute~$q_i^*$ using Algorithm~\ref{alg:Pheromone_Combine}
                        \Else
                            \State Exploitation Mode control: Compute~$q_i^*$ using~\eqref{eqn:Mode2_Waypoint}
                        \EndIf
                        \State Compute input~$u_i(t)$ to set $q_i^* \to 0$
                    \EndFor
                \end{algorithmic}
            \end{algorithm}

            The fully synthesized algorithm is given in Algorithm~\ref{alg:AgentControl_Full}. Each processing step of the agent starts with agent broadcasting its local pheromone list and local target list, subsequently receiving the same from neighbors along with the relative position of its neighbors. The agent then senses the targets in FOV and gets the state estimates for them. The agents then update the pheromone storage (Algorithm \ref{alg:Pheromone_DataStorage}) based on the received pheromone list from neighbors and local displacement data. Simultaneously, it also updates the target storage (Algorithm \ref{alg:TargetStorage}) based on the local sensing data and received target list from the neighbors. It will then attempt to choose a target based on the data in the local target list and the neighbor target list using Algorithm \ref{alg:TargetChoosing}. If a valid target is chosen, then the relative waypoint to that target is set using~\eqref{eqn:Mode2_Waypoint}. If there are no targets chosen, then the agent will execute the pheromone based searching as seen in Algorithm \ref{alg:Pheromone_Combine} and choose the appropriate waypoint. Finally, the agent will attempt to navigate to this relative waypoint using the control law~$u_i(t)$. 

        \subsection{Memory and Computational complexity}
            The storage structure of the algorithm increases linearly with the maximum number of pheromones, maximum number of targets and the maximum number of neighbors. Hence, the memory complexity is $O(N_s + N_h + \pherweightinit / \pherweightdecay)$
            As for the computational complexity, for the virtual pheromone subsystem, we have a linear time complexity based on the maximum number of pheromones and the number of neighbors. Hence, it's $O(N_s + \pherweightinit / \pherweightdecay)$ for pheromone subsystem. For the target tracking subsystem, however, the complexity for comparison of entropy of the targets is linear in the number of targets, but increases by a squared factor for maximum number of neighbors. Hence, it's $O({N_s}^2 + N_h)$ for the target subsystem. The overall computational complexity of the proposed solution thus becomes $O({N_s}^2 + N_h + \pherweightinit / \pherweightdecay)$.

        \begin{remark} [Collision avoidance]
            \rm{We note that collision avoidance is not considered in the algorithm. For the pheromone based search strategy, the goal is to maximize the effective search area and this leads to agents naturally attempting to spread out. For target tracking, our strategy relies on only one agent being assigned to a found target. The agents without any targets found will follow the search strategy. With a natural tendency for agents to spread out, collisions between agents can be completely avoided by adding any existing collision avoidance strategies like ORCA \cite{Berg2011} with minimal impact to the search performance. We thus are not formally discussing collision avoidance in this paper.}
        \end{remark}

%%%%%%%%%%%%%%%%%%%%%%%%%%%%%%%%%%%%%%%%%%
    \section{Ideal Simulation in 2D}
    \label{sed:simulation}
        
        Consider a set of agents with unicycle dynamics~\eqref{eqn:AgentDynamics_Unicycle} and a set of targets which move in a Brownian motion in a 2D space of size $30 \unit{bl} \times 30 \unit{bl}$. Here, $\unit{bl}$ denotes the body length of the agent. The dynamics of each agent~$i$ is        
        
        \begin{align}
            \label{eqn:AgentDynamics_Unicycle}
            p_i(t + 1) = p_i(t) + \begin{bmatrix}
                u_i^1(t) \cos p^\theta_i(t) \\
                u_i^1(t) \sin p^\theta_i(t) \\
                u_i^2(t)
            \end{bmatrix} , \; |u_i(t)| \leq \bar{U}_i,
        \end{align}

        where~$p_i(t) = [p_i^x, p_i^y, p_i^\theta]^T$ and the inputs $u_i(t) = [u_i^1(t), u_i^2(t)]^T$ are bounded by $\bar{U}_i$. In this simulation, we assume the bounds on input as $\bar{U}_i = [0.4\unit{bl \per \s}, 15\unit{\degree \per \s}]^T$. 

        The sensing suite on the agents is modelled after a camera system. The measurement from the sensor is the reading of the relative bearing and the relative range of the target. Hence, the measurement model is

        \begin{align}
            \label{eqn:CameraDetectionModel}
            \begin{aligned}    
                \hat{q}_{i, k}^{\text{pol}} &= \begin{cases}
                    q_{i,k}^{\text{pol}}(t) + v_{ik}(t) \quad \text{if} \; x_k(t) \in \text{FOV}_i(t), \\
                    \emptyset, \quad \text{otherwise},
                \end{cases}  
            \end{aligned}
        \end{align}

        where $q_{i,k}^{\text{pol}}(t) := [q_{i,k}^r, q_{i,k}^\phi]^T$ is the relative polar coordinates of the target. 
          
\begin{figure}[t]
            \centering
            \begin{subfigure}{0.5\linewidth}
                \centering
                \includegraphics[width=0.7\linewidth]{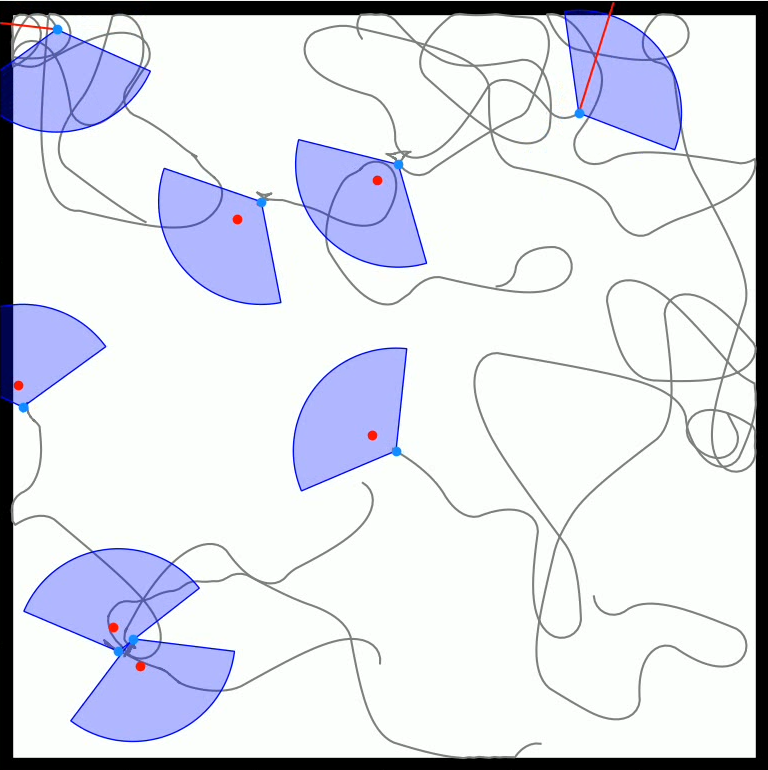}
                \subcaption{Environment Map}
                \label{fig:Sim_EnvironmentPlot}
            \end{subfigure}
            \hfill
            \begin{subfigure}{\linewidth}
                \centering
                \includegraphics[width=0.9\linewidth]{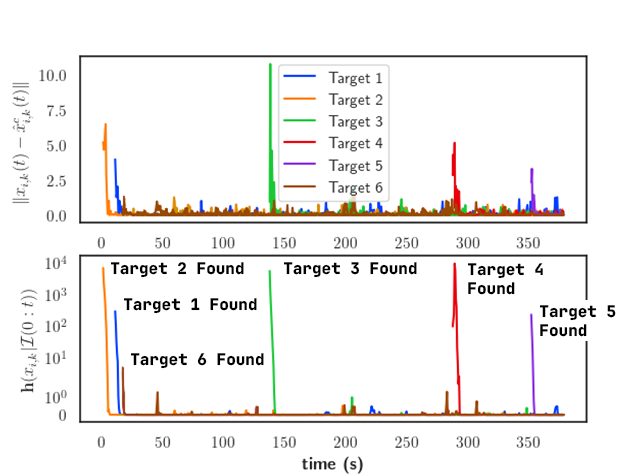}
                \subcaption{Estimation error}
                \label{fig:sim_h_e_plots}
            \end{subfigure}
            \caption{Simulation of the active perception algorithm. The environment map is shown in (a) with agents in blue, FOVs in purple, targets in red, waypoints being the red lines and agent trajectories being the gray lines. In (b), we plot the $\TwoNorm{x_{i, k}(t) - \hat{x}_{i, k}^c(t)}$ and $\mathbf{h}(x_{i, k}|\mathcal{I}(0:t))$ as time passes.} %At the right is the averages of time to initiate tracking $n$ targets simultaneously shown on the top with the number of simulations which had valid values in the bottom.}
            \label{fig:SimResults}
        \end{figure}

        The FOV of the sensor is defined as a sector centered at the current position of the agent. We show an illustration in Figure \ref{fig:Sim_EnvironmentPlot}. Here we have the sensing radius $r_s = 4\unit{bl}$ and the sector angle of the sensing region $\phi_c = 120 \unit{\degree}$.

        We compute the relative position in cartesian coordinates as the estimate of the target $\hat{x}_{i, k}^{\text{raw}}$ and obtain the corresponding covariance $\Sigma_{i, k}(t)$ from the covariance map $R_i^s(q_{i, k})$. We show the exact formulation of the raw target estimates and the covariance map used below:

        \begin{subequations}
            \label{eqn:Estimate_CameraSenseToRelEstimate}
            \begin{align}
                \hat{x}_{i, k}^{\text{raw}} &= 
                    \begin{bmatrix}
                        \hat{q}_{i,k}^r \cos \hat{q}_{i,k}^\theta\\
                        \hat{q}_{i,k}^r \sin \hat{q}_{i,k}^\theta
                    \end{bmatrix},  \\
                R_i^s(\hat{q}_{i, k}) &= \mathbf{T}_r \begin{bmatrix}
                    \eta_i(\hat{q}_{i, k}) & 0\\
                    0 &  \eta_i(\hat{q}_{i, k})
                \end{bmatrix} \mathbf{T}_r^T, \\
                \mathbf{T}_r &=
                    \begin{bmatrix}
                        \cos \hat{q}_{i,k}^\theta & -\sin \hat{q}_{i,k}^\theta\\
                        \sin \hat{q}_{i,k}^\theta & \cos \hat{q}_{i,k}^\theta
                    \end{bmatrix} , \\
                \eta_i(\hat{q}_{i, k}) &= k_1^s (\hat{q}_{i,k}^r(t) - \bar{r})^2 + k_2^s \hat{q}_{i,k}^\phi(t)^4,
            \end{align}
        \end{subequations}
          where $\bar{r}$ denotes the distance from the sensor where we get the best estimate. Further, the covariance map for other-agent sensing $\neighposcovmap = \diag {k^p || p_i - p_j ||}$. For the simulations, we choose $k_1^s = 1, k_2^s = 1$ and $k^p = 1$.

        The communication distance $r_c$ for the simulations is $12 \unit{bl}$. Each agent receives data only once every three processing steps. The initial pheromone strength is $w_i^d = 35$, the pheromone decay is $\pherweightdecay = 0.16$ and the pheromone threshold $\pherweightthresh = 0.1$. A lower $w_i^d / \pherweightdecay$ yields a shorter pheromone history resulting in more frequent revisit of locations and vice-versa. Once the agent computes a waypoint, the agents move towards the waypoints using a very simple PD controller that attempts to get the agent to the relative waypoint computed $q_i^*$. 
            
        To test the performance, we simulated $N_s=8$ agents and $N_h=6$ targets with random initial positions in the environment. The agents have no prior knowledge of the targets. The result of one of the runs is shown in Figure~\ref{fig:Sim_EnvironmentPlot}. The proposed algorithm allows the agents to detect all 6 targets in the environment and keep track of them. To show the feasibility of the parameter $\mathbf{h}(x_{i, k}|\mathcal{I}(0:t))$, we plot the observed value in the simulation against the actual error of estimation $\TwoNorm{x_{i, k}(t) - \hat{x}_{i, k}^c(t)}$ shown in Figure \ref{fig:sim_h_e_plots}. We see that each discovery of a target comes with a high value for $\mathbf{h}(x_{i, k}|\mathcal{I}(0:t))$. However, the values quickly drops to close to zero as the estimation gets better. Using the simulation results, we can conclude that keeping track of the measurable quantity $\mathbf{h}(x_{i, k}|\mathcal{I}(0:t))$ allows the agent to determine how good the estimate will be and move to a desirable relative position that attempts to minimize the estimation errors.  

        We performed 60 simulations where the randomly initialized agents use the proposed searching and target tracking algorithms with the above-mentioned specifications to seek randomly initialized targets.

        \begin{table} [h]
            \centering
            \scalebox{0.78}{
            \begin{tabular}{|c|c|c|c|}
                \hline
                Algorithm & Avg. time to track & N.Simulations & Complexity \\
                \hline
                \hline
                Auction & 614.1 & 60 & $O((N_s)^3)$\\
                \bf{Proposed} & 640.5 & 60 & $O((N_s)^2 + N_h)$ \\
                Local Greedy & 343.6 & 18 & $O(N_h)$\\
                \hline
            \end{tabular}}
            \caption{Statistics from 60 simulations for time to simultaneously track all 4 targets in a simulation consisting of 6 agents which compares the effects of target assignment algorithm. Here N.Simulations denote the number of simulations where all 4 targets were tracked. All agents are searching using the pheromone (proposed) search algorithm.}
            \label{tbl:Time2Track_nTargetSelAlgs}
        \end{table}

        For comparing target selection, the baseline algorithm used for comparison is ``local greedy'', where each agent tracks the target with the least uncertainty in its own FOV. We also compare target assignment against the popular auction algorithm~\cite{bertsekas1979} as a more capable algorithm, albeit with higher compute requirements. The detection and actuation capabilities of all agents in these simulations are the same. In all cases we use the time to track targets as the metric to compare performances. The comparison between our proposed target selection strategy and the popular auction algorithm yields no significant performance difference between them as evidenced by table~\ref{tbl:Time2Track_nTargetSelAlgs}. With the auction algorithm being more computationally expensive, the choice of a simpler distributed greedy strategy is a more scalable option. Comparing the proposed target assignment strategy against a ``local greedy'' assignment, there were very few simulations where all targets were tracked as shown in table~\ref{tbl:Time2Track_nTargetSelAlgs} and this proves to be an undesirable behavior. The proposed algorithm, in contrast, finds all targets in the environment. 

        \begin{table}[h]
            \centering
            \scalebox{0.85}{
            \begin{tabular}{|c|c|c|}
                \hline
                & \multicolumn{2}{c|}{Requirements} \\
                Search Algorithm & Map Bounds & Agent localization \\
                \hline
                Anti-Flocking & Required & Global \\
                \bf{Pheromone (Ours)} & Not Required & Relative \\
                Levy & Not Required & Not Required \\
                \hline
            \end{tabular}}
            \caption{Comparison of sensing requirement for different search algorithms}
            \label{tbl:searchAlgReq}
        \end{table}

        % Search strategy
        \begin{figure}[h]
            \centering
            \includegraphics[width=0.9\linewidth]{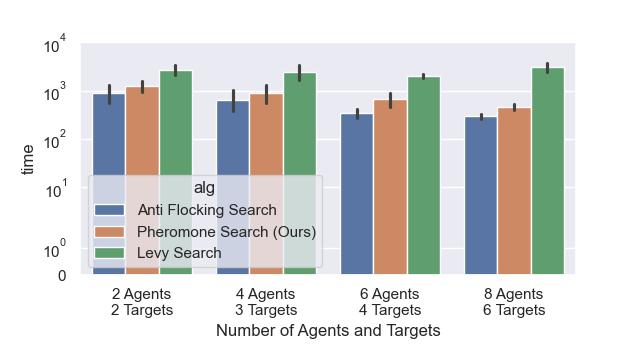}
            \caption{Statistics of 60 simulations for comparison of search algorithms where the time to track all targets in the environment for different numbers of agents and targets is explored. The environment size is $30\unit{bl} \times 30\unit{bl}$.}
        \end{figure}

        To compare the performance of the proposed search algorithm, we use the Levy's walk~\cite{Viswanathan1999, Sutantyo2010} as a baseline searching algorithm with lesser sensing and communication capabilities when searching for targets. In~\cite{Zedadra2019}, we see that Levy walk is mentioned as one of the most efficient algorithms for searching when targets are sparse in the environment. We also study the search performance against the more capable Anti-Flocking algorithm~\cite{Ganganath2016}, where all agents require a GPS sensing modality. The sensing requirements of these algorithms are summarized in table~\ref{tbl:searchAlgReq}. We look at the time to track statistics for all targets. Comparing our proposed search algorithm to levy walk, we see that there is much smaller performance gains when the number of agents are small. The performance gains increase exponentially with the number of agents. We see a maximum of 4x improvement for the case with 8 agents and 6 targets in the environment. We can clearly see the effects of coordination between the agents here improving the search and tracking performance with the number of agents. When comparing against the more capable Anti-Flocking algorithm, we see about $30\%$ more time to track all targets. Since our proposed solution does not use GPS for the search strategy, the loss in performance can be attributed to that. 

        \begin{figure}[h]
            \centering
            \includegraphics[width=0.85\linewidth]{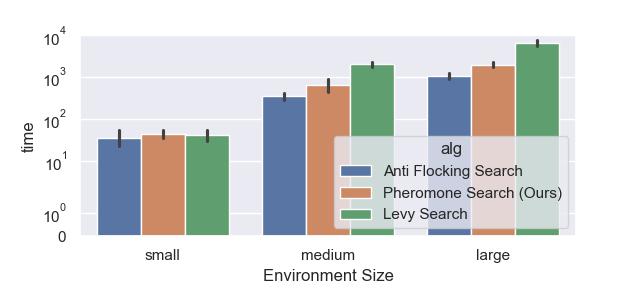}
            \caption{Time to Track Statistics of 60 simulations for varying sizes of the environment for a 6 agent and 4 target simulations. Small environments are $10\unit{bl} \times 10\unit{bl}$, medium environments are $30\unit{bl} \times 30\unit{bl}$ and large environments are $50\unit{bl} \times 50\unit{bl}$.}
            \label{fig:Time2Track_envSize}
        \end{figure}
        
        \begin{figure}[h]
            \centering
            \includegraphics[width=0.85\linewidth]{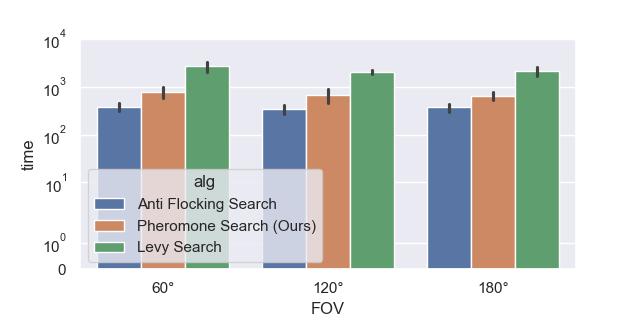}
            \caption{Time to Track Statistics of 60 simulations for varying sizes of the agent Field of View. Here there are 6 agents and 4 targets; The environment size is $30\unit{bl} \times 30\unit{bl}$.}
            \label{fig:Time2Track_FOV}
        \end{figure}
        
        We also study the performance of the search strategy in varying size of the environment and the FOV.  Comparing the time to track performance by varying the environment sizes as seen in Figure~\ref{fig:Time2Track_envSize}, we see that in smaller environments ($10\unit{bl} \times 10\unit{bl}$), levy walk is doing slightly better than our proposed algorithm. This is an interesting observation and seems to point to the fact that at higher agent densities, the pheromone based search provides very little improvement. The medium ($30\unit{bl} \times 30\unit{bl}$) and the large ($50\unit{bl} \times 50\unit{bl}$) environments have similar performance gains over levy walk as expected. Finally, comparing different FOVs, we see very similar performance in all simulations. Hence, in environments with large number of agents where the agent density is not very high, our proposed solution performs well over the baseline strategy for the problem of online distributed active perception with limited information.

    \section{Real Robot Experiments}
    \label{sec:toRealRobots}
    
        Here we consider a scenario in a which an autonomous Lighter-Than-Air (LTA) multi-robot system must be deployed to find and track neutrally buoyant balloons in an unknown environment. In order to apply the proposed distributed active perception algorithm on real robotic system,  we show a specific hardware instantiation that solves the Problem~\ref{pr:activeperception} in a target tracking scenario.

        \begin{figure}[ht]
            \centering
            \includegraphics[width=0.7\linewidth]{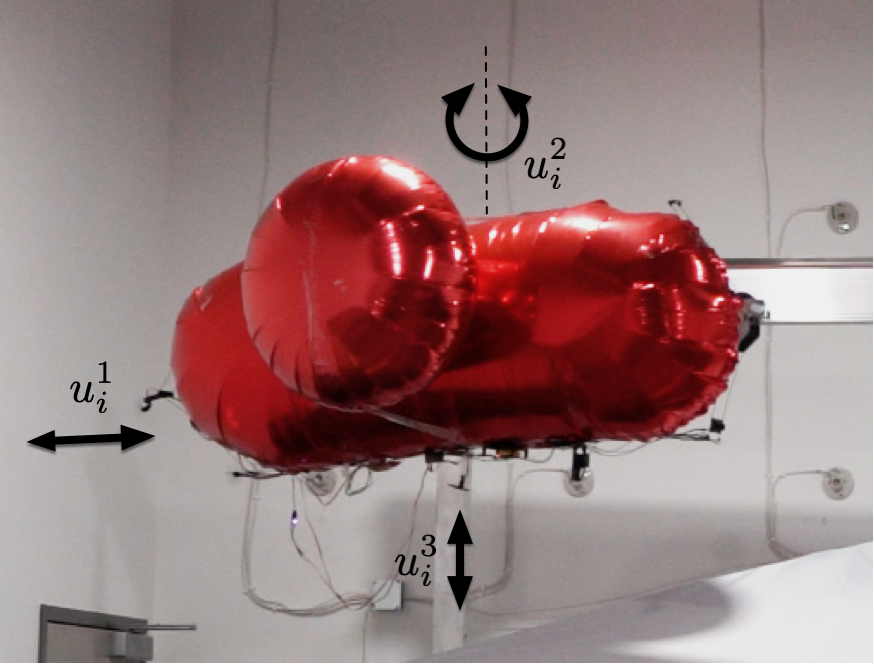}
            \caption{Lighter than Air (LTA) agent showing the control inputs available in the system.}
            \label{fig:LTAAgent}
        \end{figure}

        We model our LTA agent shown in Figure~\ref{fig:LTAAgent} as a kinematic rigid body in 3D space with position 
        \begin{align*}
        p_i = \begin{bmatrix} p_i^x & p_i^y & p_i^z & p_i^\theta & p_i^\phi & p_i^\psi \end{bmatrix}^T .
        \end{align*}
        The chosen actuator configuration shown in Figure~\ref{fig:LTAAgent} allows us to independently move the agent forward/backward using~$u_i^1$, turn clockwise/counter-clockwise using~$u_i^2$ and move up/down using~$u_i^3$. 

        Based on our agent design and actuator configuration, we assume that the agent does not pitch or roll~(i.e.,~$p_i^\phi = p_i^\psi = 0$ at all times). The reduced-order kinematics we consider for controlling our robots are
                \begin{align}\label{eq:realdynamics}
                    p_i(t + 1) = p_i(t) + \begin{bmatrix}
                        u_i^1(t) \cos p^\theta_i(t) \\
                        u_i^1(t) \sin p^\theta_i(t) \\
                        u_i^3(t) \\
                        u_i^2(t) \\ 
                        0 \\ 
                        0
                    \end{bmatrix} , \; u_i(t) \leq \bar{U}_i.
                \end{align}
        
        The targets are neutrally buoyant green balls. They can freely move in the space in the XY plane. There is a slight restriction in the z axis motion by means of a string attached to the ball with a small weight on the end.  Using very simple experiment of capturing the time varying motion of the balls in the environment and estimating the maximum motion of the target, we can estimate the upper bound on the target noise $\bar{Q}_k$ using gaussian assumptions.
        
        \begin{remark} [Sim. - Expm. Compatibility]
            \rm{Compared to the agent discussed in simulation (2D unicycle model), we see that this real robotic system has an additional independent z axis input. We can consider this to be a 2.5D system instead of a full 3D system owing to the zero pitch and roll. Using a combination of instantiation of the targets restricted to same height (using a string with an attached weight that can move on ground) and limiting the motor control commands for $u_i^3(t)$, we assume here that the agent will mostly move with much smaller Z height variation as compared to its translation in XY plane. Thus, we expect to see parity between the behavior shown in the 2D simulation and the 2.5D LTA agents.
            }
        \end{remark}
        
        We now show exactly how our agent satisfies the minimum capabilities required to implement this algorithm and how we realize Assumptions~\ref{asn:MaxVelocity}-\ref{asn:communication}.

        \subsection{Actuation, Communication, Sensing, and Perception}

            Given our actuator configuration, we can easily satisfy Assumption~\ref{asn:WaypointValidity} using three independent PD controllers in parallel to move towards a desired relative waypoint~$q_i^*$. Inputs~$u_i^2$ and~$u_i^3$ independently control the height and yaw of the agent, while~$u_i^1$ controls for forward motion.
        
            To enable a method of agent-to-agent broadcast communication we have developed a communication system called ReLoki that sends out the local pheromone $L_i^p$ and target lists $L_i^s$ as formatted 802.15.4 data packets. The system uses a UWB based DW1000 module as the transceiver in both the sending and receiving agent. We call this system ReLoki, and it is capable of allowing agents to communicate satisfying Assumption~\ref{asn:communication} with a distance~$r_c = 6.5\unit{m}$. In Section~\ref{se:relativeposition_hardware}, we discuss the communication protocol used.

            To implement the proposed target searching and tracking algorithm proposed in Section~\ref{sec:DistributedActivePerception}, we must develop and fully integrate an end-to-end system capable of providing the minimum level of sensing mentioned in Section~\ref{se:sense}. The system utilizes a camera based target detection system that uses a pre-trained object detection frame-work for obtaining the raw target observations $\targetsensraw$. To get the relative position $\neighpos$ between agents, the system uses the previously mentioned ReLoki module. With each communication transaction between agents, the message will contain the relative position of the neighboring agent. Finally, the displacement sensing $\robotposlast$ is done using an onboard IMU performing dead-reckoning. The full details of this are available in Appendix. 

    \subsection{Hardware Experiments}

    \begin{figure}[t]
            \centering
            \includegraphics[width=0.5\linewidth]{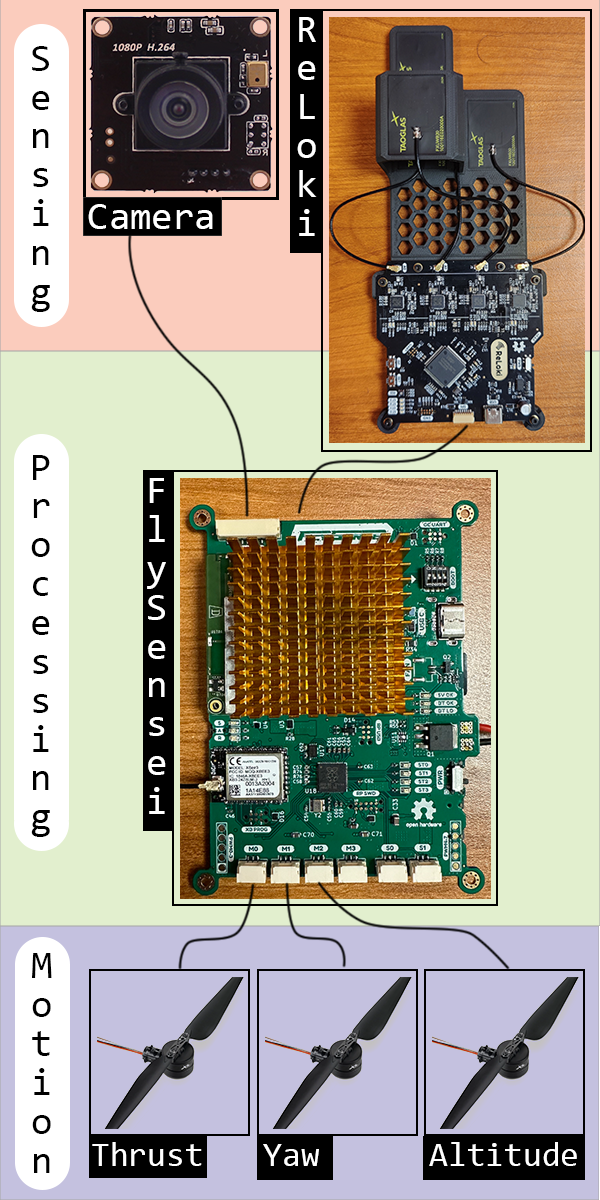}
            \caption{Complete onboard system attached to the blimps to perform the search and track. Here the vision processing and displacement processing is performed by FlySensei and the inter-agent position sensing and communication is performed by ReLoki. The Algorithm runs on FlySensei generating motor commands for the blimps.}
            \label{fig:FullSystem}
        \end{figure}
    
	Now with methods in place to compute~$\targetsensraw$, $\neighpos$ and~$\robotposlast$, we show how Algorithm~\ref{alg:AgentControl_Full} can be realized on our LTA robots. We use two collaborating LTA agents for this experiment. These LTA agents have the sensing capabilities mentioned in section \ref{sec:toRealRobots}. We designed a controller called FlySensei that houses the target detection system, the displacement sensing system and other systems required for the actuation of agents. ReLoki then interfaces with this main controller. All the processing required for the proposed search and track algorithm runs on the FlySensei system which eventually generates the motor control commands required to move the LTA agents in the environment. We show an illustration of the sensing, processing and actuation system in Figure~\ref{fig:FullSystem}. 

    The testing area is an enclosed room of dimensions $10 \times 6 \unit{m}$ with the LTA agents initialized at the center of the room with no view of the targets. The targets to detect are a purple ball and green ball scattered at two ends of the environment. The parameters used for testing is shown in Table~\ref{tbl:ExpmParameters}. Please note here that the parameters with ``Expm.'' are lookup table values that are generated from the covariance map generation as mentioned in the Appendix. 

    \begin{table} [ht]
        \centering
        \scalebox{0.78}{
        \begin{tabular}{||c|c|c||}
            \hline
             Variable & Description &Value \\
            \hline \hline
            $v^p_\text{max}$            & Agent Maximum Speed               & 0.6 \unit{m/s} \\
            $r_s$                       & Sensing FOV range                 & 5.5 \unit{m }\\
            $\phi_s$                    & Sensing FOV Bearing Angle         & 120 \unit{\degree} \\
            $R_i^s(.)$                  & Camera System Covariance Map      & Expm.    \\
            $\bar{Q}_k$                 & Target Covariance Growth          & \text{diag}[0.19 \:	0.15 \: 0.28]      \\
            $\bar{\Sigma}_i$            & Target Max Covariance             & 3600     \\
            $R^p_{i, j}(.)$             & ReLoki Covariance Map             & Expm     \\
            $r_c$                       & Communication range               & 6.5m      \\
            $w_i^\text{init}$           & Initial Pheromone Weight          & 15      \\
            $\pherweightdecay$          & Pheromone Decay                   & 0.3      \\
            $\pherweightthresh$         & Pheromone Threshold               & 0.1      \\
            $\robotposlastcovmap$       & Displacement covariance map       & Expm      \\
            \hline
        \end{tabular}}
        \caption{Parameters used for experiment.}
        \label{tbl:ExpmParameters}
    \end{table}

\begin{figure}[ht]
        \centering
        \includegraphics[width=\linewidth]{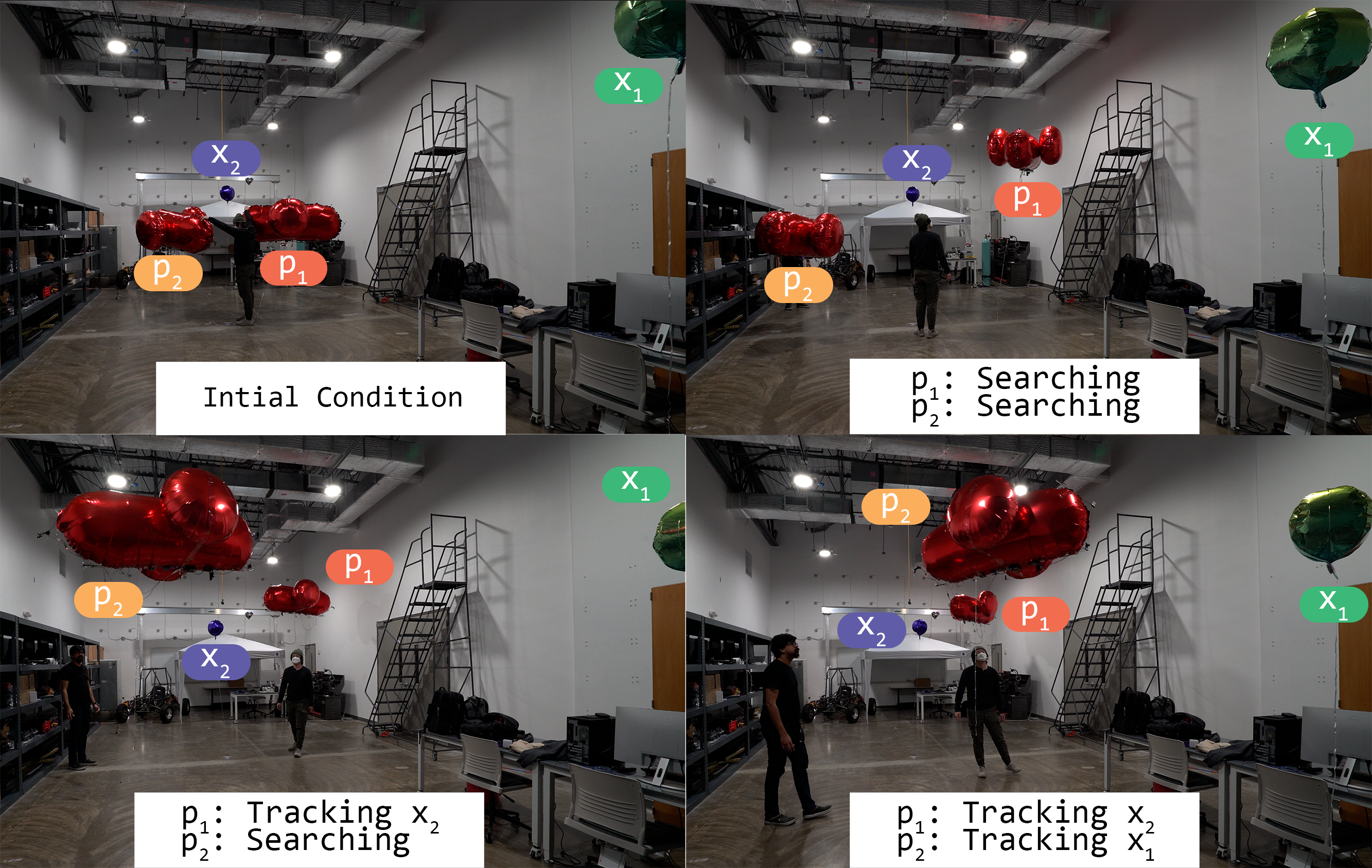}
        \caption{Experiment progression in time showing key moments of searching and tracking 2 targets by 2 blimps.}
        \label{fig:ExpmFlow}
    \end{figure}

    We show the progression of one of the experimental runs where the system tracks 2 targets in Figure~\ref{fig:ExpmFlow}. We have observed from all the experimental runs that they quickly try to move to different parts of the room and search for targets. Once they find targets, we observed that in most of the cases they approach and keep tracking the target. There were however a few experiments where they did not keep the target lock. In these cases, the agent switched back to search and then subsequently it found and traced the target. We attribute these cases to a combination of the PID tuning and the random disturbances due to sudden gusts in the room. 

\begin{figure}[h]
        \centering
        \includegraphics[width=0.9\linewidth]{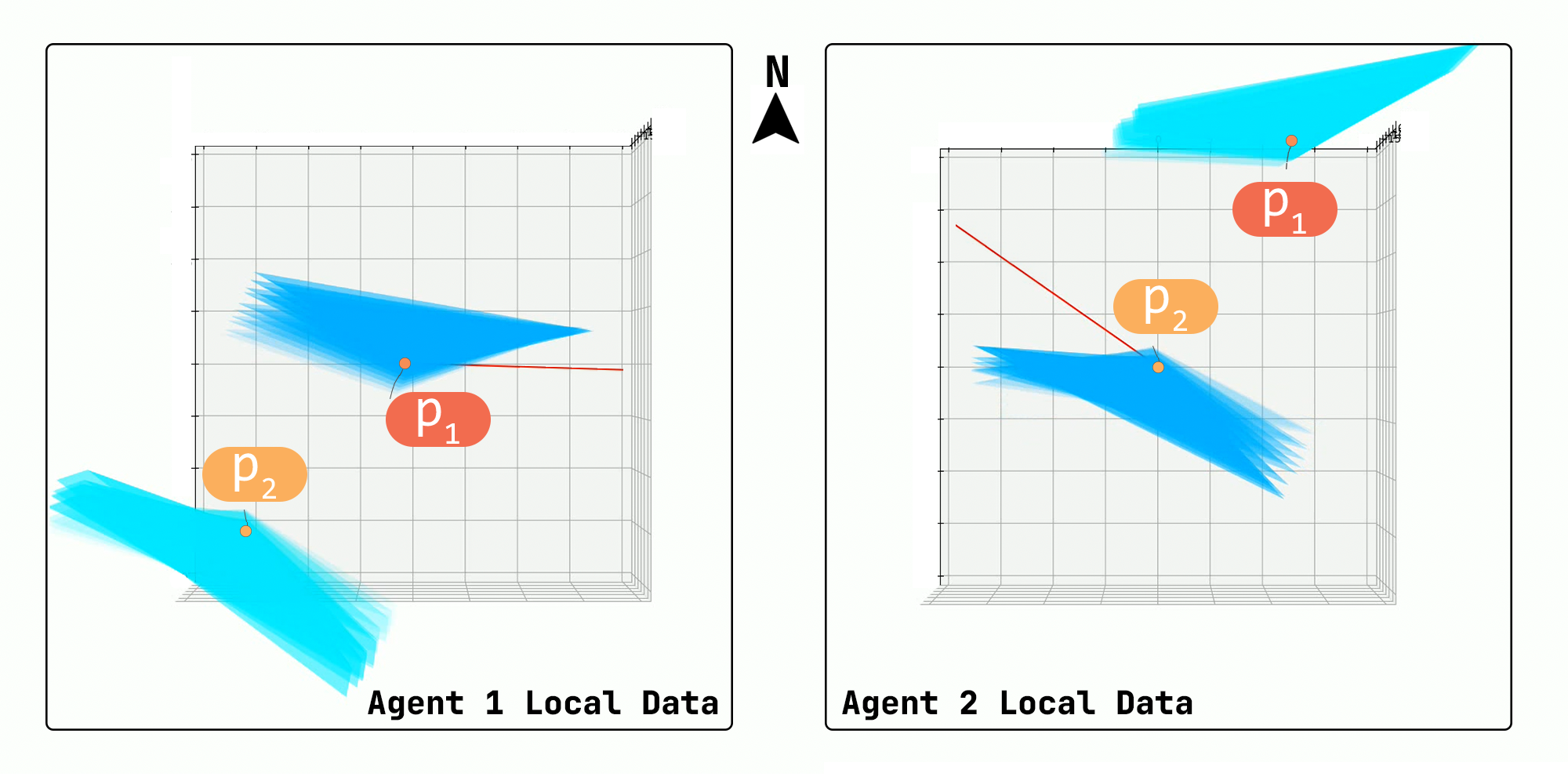}
        \caption{Pheromone Map generated by the agents used in the experiment with Agent 1 pheromone map on the left and agent 2 pheromone map on the right. In each image, the deep blue region shows the pheromone map of the local agent and the cyan shows the pheromone map of the neighboring agent. The intensity of the color translates to the pheromone weight. The red line shows the waypoint chosen by the agent. Both images are oriented with north facing up.}
        \label{fig:ExpmCovariacnePlot}
    \end{figure}

    To show that the searching strategy of the system works as seen in the simulations, we visualize one time instance of the pheromone map \eqref{eqn:PheromoneMap} created by both agents in Figure \ref{fig:ExpmCovariacnePlot}. Here we see that agent 1 has the latest information about agent 2, however agent 2 has a slightly older information. This is to be expected as the communication between agents happen only every 2 seconds. We see that the algorithm clearly makes waypoint choice that will move the agents to unexplored areas i.e. areas with the least pheromones. This is translated to the agents moving to different corners of the room as observed in Figure~\ref{fig:ExpmFlow}. We do see some parity between the simulation and the real robots here. However, due to the drifts happening in the IMU based displacement sensing, we have observed that the search is not as effective as seen in simulations. 

\begin{figure}[h]
        \centering
        \includegraphics[width=0.9\linewidth]{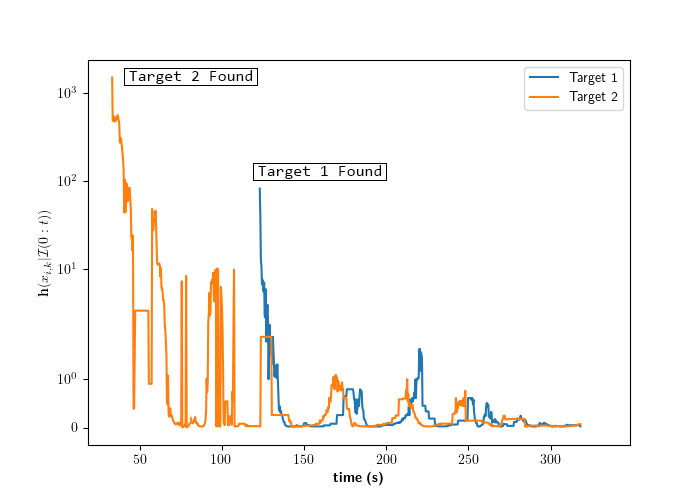}
        \caption{Entropy of targets from agents. Target 2 is found first by agent 1, and it is tracked until experiment end. Similarly, target 1 is found later by agent 1. We stop the experiment when both targets are tracked for some time.}
        \label{fig:ExpmEntropy}
    \end{figure}

    We then look at the target capture. The entropy of the target as measured by the agent (similarly to that of simulations) is shown in Figure~\ref{fig:ExpmEntropy}. We observe that once the targets are found in this case, the agent keeps track of the target until experiment end. Looking at the entropy plot, even though we see the trend of the entropy going lower, there are areas where the entropy increases locally. We attribute this to both the irregular covariance map of the sensing system and the PID based tracking system having slight oscillations. These are the kind of inconsistencies that were not observed in the ideal simulation. We also tested the target negotiation between the agents by initializing 2 agents looking at the same target and have found that when both agents have target in FOV, the agent with the lower entropy of the target seems to select the target while the other moves along searching for other balls.

    \section{Conclusion}
    In this paper we present a system integration based solution for a real-time decentralized active perception problem, where the agents only communicate using broadcast messages and uses no global positioning data. We decomposed the problem to two specific sub-problems. One of them deals with dynamic coverage control and uses a virtual pheromone based solution where the agents attempt to maximize the probability of finding targets. The second is a decentralized target assignment problem that enables the agents to negotiate which targets to track. Each agent chooses to track one target and moves itself to a relative position that has the least detection uncertainty. Both these solutions work in tandem to find targets and reduce the system-level uncertainties about the targets of interest. We test the proposed solution against base-line search strategy, which shows 4x performance improvement.  We also developed the required sensing modalities to enable the proposed searching and tracking system and in a case study show the agents tracking free moving balls instantiated at previously unknown locations. 
	
    %In the future we plan to improve the performance of the system by using a simulation based performance analysis for potential upgrades to the existing system and implementing the same in LTA Agents. We envision this system being useful in search and rescue scenarios by augmenting first-responders with robotic agents performing the proposed algorithm and eventually making this a "deploy and forget" system. 

    \section*{Acknowledgements}
    This work was supported in part by the Department of the Navy, Office of Naval Research (ONR), under federal grant N00014-20-1-2507. We also acknowledge Kevin Zhu for his invaluable contributions to the design of the LTA agent and his work in experimental setup of the robotic agents. We also acknowledge Kentaro Nojima-Schmunk for the helium filled envelopes he developed for the experiments. 

    \bibliographystyle{resources/bst/sn-chicago}
    \bibliography{resources/References.bib}

    \begin{appendices}

    \section{Sensing and Perception Implementation}
        \subsection{Target Detection System}\label{se:detection_hardware}
            To detect a target in the vicinity of the agents, one of the ubiquitous methods is using a camera. The data can be processed to obtain relative location of the target as long as the target is in the FOV of the sensor. We have seen a lot of object detection systems gaining prominence in the last decade~\cite{Zou2019,Zhao2019,Redmon2016,Girshick2015} with advanced image segmentation techniques~\cite{Minaee2022} that allow even 3D pose estimation in some scenarios~\cite{Hu2019}. We will be focusing on a simpler object detection framework in this proof of concept.
            
\begin{figure}[h]
                \centering
                \includegraphics[width=0.7\linewidth]{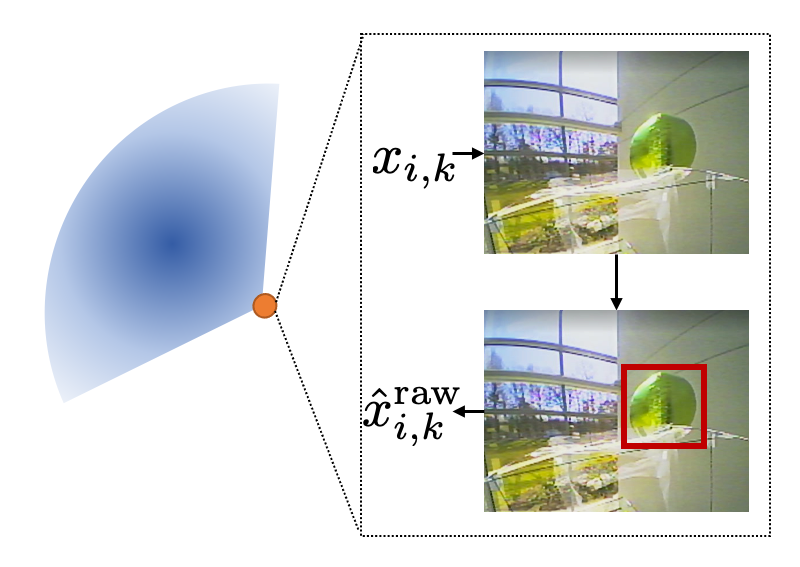}
                \caption{Illustration of the bounding box around the detected target in the FOV of the agent. The bounding box is used to extract the relative position of the target from the agent.}
                \label{fig:ObjectDetectionCamera}
            \end{figure}

            Here we show how our hardware and software enables us to generate estimates~\eqref{eqn:HiddenStateMeasurement_Sensing} by finding the appropriate mapping~$\mathbf{D}_k : \real^{d_p} \times \real^{d_h} \to \real^{d_s}$ when a green ball is in the FOV of the robot as shown in Figure~\ref{fig:ObjectDetectionCamera}.

            In our system we use an Omnivision OV5645 camera attached in front of the robotic agent that pipes visual data to a Google coral SOM to process this visual data. To detect objects we deploy a pre-trained yolov5~\cite{Jocher2022} object detection model. We have shown a generic camera model in~\eqref{eqn:CameraDetectionModel} and here we extend this to the output received from the detection system. Yolov5 outputs a bounding box around the detected target as shown in Figure~\ref{fig:ObjectDetectionCamera}. Using the bounding box, we can extract the relative position of the target~$\hat{q}_{i, k}^{\text{pol}} \in \real^3 := [\hat{q}^r_{i, k}, \hat{q}^\theta_{i, k}, \hat{q}^\psi_{i, k}]^T$ in polar coordinates. 

            \subsubsection{Target State transformation from bounding boxes} \label{se:targetstate_bb}
            
                Let the bounding box for the target~$k$ detected by agent~$i$ be defined as~$\hat{q}^{bb}_{i, k} \in \real^4 := [\hat{q}^{bbx^-}_{i, k}, \hat{q}^{bbx^+}_{i, k}, \hat{q}^{bby^-}_{i, k}, \hat{q}^{bby^+}_{i, k}]^T$ consisting of the bounds of the box in each orthogonal axes of the frame. We can extract the relative position of the target~$\hat{q}_{i, k}^{\text{pol}} \in \real^3 := [\hat{q}^r_{i, k}, \hat{q}^\theta_{i, k}, \hat{q}^\psi_{i, k}]^T$ of the object using the bounding box using
                
                \begin{align}
                \hat{q}_{i, k}^{\text{pol}} = \begin{bmatrix}
                    \sqrt{k^r_i / \left((\hat{q}^{bbx^+}_{i, k} - \hat{q}^{bbx^-}_{i, k}) (\hat{q}^{bby^+}_{i, k} - \hat{q}^{bby^-}_{i, k})\right)}\\
                    k^\theta_i (\hat{q}^{bbx^+}_{i, k} + \hat{q}^{bbx^-}_{i, k}) / 2 \\
                    k^\psi_i (\hat{q}^{bby^+}_{i, k} + \hat{q}^{bby^-}_{i, k}) / 2
                \end{bmatrix},
                \end{align}

                where~$k^\theta_i$ and~$k^\psi_i$ is the transformation factor required to change the pixel value of the detected object to bearing angles in the horizontal and vertical axes respectively. This comes out to be the FOV angle of the camera. For the specific lens-camera combination used in our experiment, it came out to be $120\unit{\degree}$. $k^r_i$ is the transformation factor that converts the pixel area to the distance to the target from the camera frame. We obtain this from the calibration experiment for the camera. For our ball and camera system, it came out to be $0.112\unit{m}$.
                
                Using this measurement model, we can easily see that the transformation required to get the state estimate is just a conversion from spherical coordinates to euclidean coordinates with a slight modification as seen in~\eqref{eqn:CameraSph2Euc}. Once we have the raw state estimates, we can use~\eqref{eqn:TargetMeasurement_FOV} to get the target estimates $\hat{x}_{i, k}$ used by the algorithm.
                \begin{align}
                    \label{eqn:CameraSph2Euc}
                    \hat{x}_{i, k}^{\text{raw}} (t) = \begin{bmatrix}
                        \hat{q}^r_{i, k} \cos \hat{q}^\theta_{i, k} \sin \hat{q}^\psi_{i, k} \\
                        \hat{q}^r_{i, k} \sin \hat{q}^\theta_{i, k} \sin \hat{q}^\psi_{i, k} \\
                        \hat{q}^r_{i, k} \cos \hat{q}^\psi_{i, k}
                    \end{bmatrix}
                \end{align}
                
                Any sensing modality has an associated noise, arising from the inconsistencies in measurement for a multitude of reasons. Most algorithms assume that the covariance is uniform over the entire sensing range to simplify the sensing system characteristics. However, in our proposed solution these changes in uncertainties over the sensing domain is used to make control decisions. To incorporate this noise with the active perception algorithm, we must first get the covariance map of the above-mentioned sensor~$R_i^s(.)$. 

            \subsubsection{Covariance Calibration for Detection System}\label{se:detectiondetails}
                Since we are making use of the covariance map, we are proposing an experimental setup to find that covariance map. With the target detection system developed outputting the raw estimates~$\hat{x}_{i, k}^{\text{raw}}$, we set up the system on a pan and tilt experimental contraption. We have one target in the experiment that the camera has to detect set at a fixed distance. The pan tilt mechanism allows the camera to test the detection in all possible relative positions. In the experiment we discretized the pan and tilt range in increments of $10^\circ$. We have observed that the FOV for this camera setup extends to~$120^\circ$ in the horizontal axis and vertical axes, which are our range limits. The distance for target is discretized at~$1\unit{m}$, starting from~$1.5\unit{m}$ with a maximum of~$5.5\unit{m}$, which was observed as the maximum detection distance. Please note that the above-mentioned values have to be tailored to this specific sensor-target pair. The camera system exhaustively goes through all the range-pan-tilt triplets and collect~$50$ samples of the relative position estimate for each range-pan-tilt triplet $q_m$. We also have the real relative position from the experiment. The initial covariance map is generated by finding the covariance for the~$50$ error measurements for the readings obtained.
        
                \begin{figure}[h!]
                    \centering
                    \includegraphics[width=0.8\linewidth]{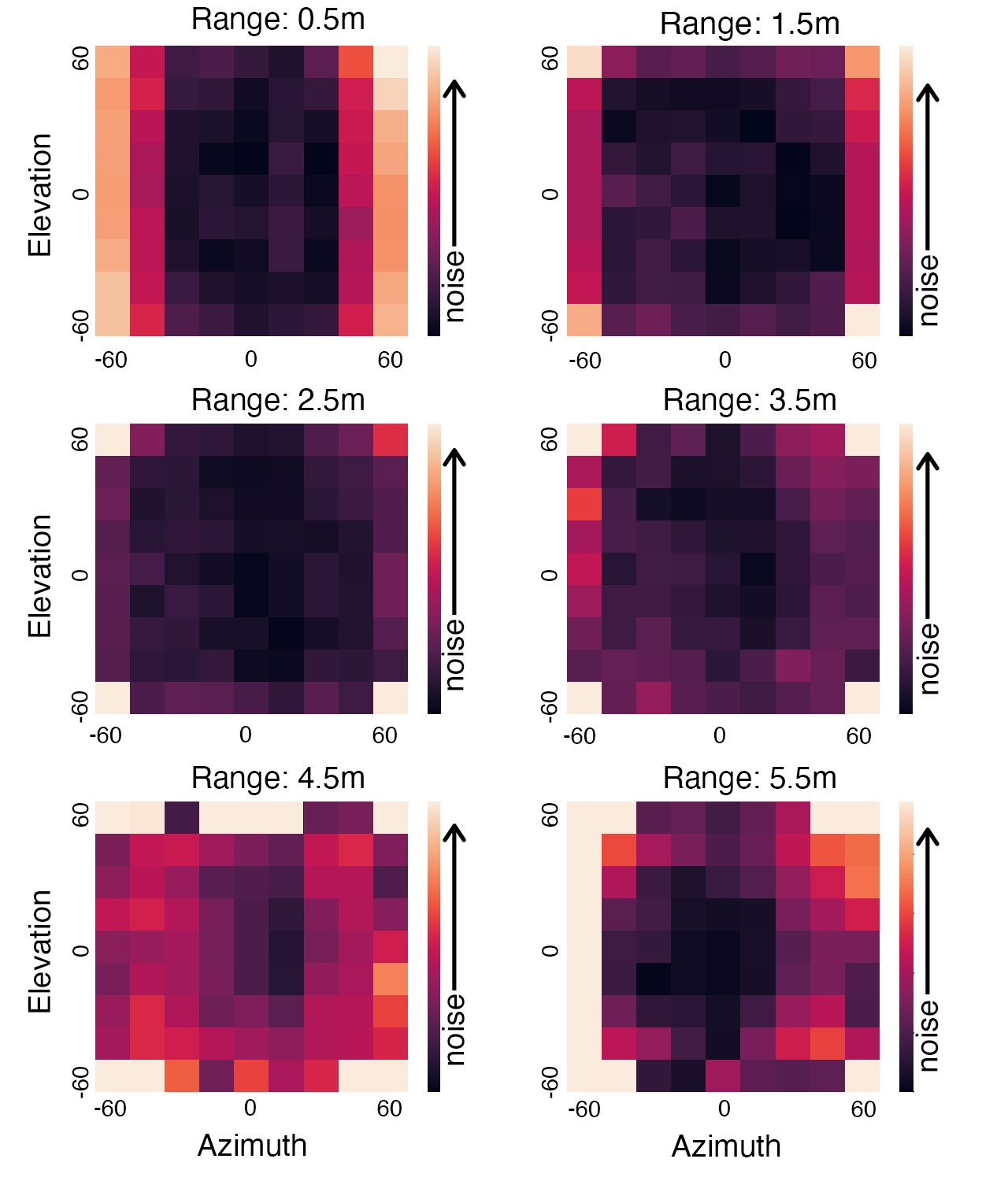}
                    \caption{Covariance map of target detection for yolov5 running on the LTA agents. Each image shows the entropy at different bearing angles for a given range.}
                    \label{fig:VisionCovarianceMap}
                \end{figure}

                This is obtained using 2 factors. First we obtain the error from the ground truth data~$\text{Cov}^e(q_m) := E[\hat{x}_{i, k}(q_m)^{\text{raw}} - x_{i, k}(q_m)]^2$ at the relative range-pan-tilt $q_m$. We also get the covariance of the measured errors $\text{Cov}^\sigma(q_m)$. The final covariance used is a very conservative measurement of the actual covariance and computed as shown in~\ref{eqn:Covariance_ConservativeExt}. This is done due to the sparse measurements taken in the sensing domain. The effect of this is somewhat mitigated using the covariance smoothing discussed later. 
        
                \begin{align}
                    \label{eqn:Covariance_ConservativeExt}
                    \text{Cov}(q_m) = \text{diag}(\text{Cov}^e(q_m)) +  \text{Cov}^\sigma(q_m).
                \end{align}
                
                The results of the covariance experiment is shown in Figure~\ref{fig:VisionCovarianceMap}. Each block shows the entropy of the measurements. We observed very good detection around $2\unit{m}$ (between $1.5\unit{m}$ and $2.5\unit{m}$) at $0^\circ$ bearing angles. This will be used as the optimal relative location to track the targets. We have observed the max range for this detection is $5.5\unit{m}$, and we were getting very spotty detection after this range. So we will be soft capping the readings to this range measurement for the full integrated hardware experiment.

                Since the map obtained in the experiment is highly discretized, to find the covariance at any arbitrary point in the FOV of the agent to be used in the sensing system, we apply an interpolation based on the observed covariance called $N$-nearest covariance interpolation 

                \begin{align}
                    R_i^s(q_{i, k}) = \frac{\sum_{\mathcal{N}_c(q_{i, k})} \left(1 - \frac{||q_{i, k} - q_m||}{r_{\text{max}}}\right) \text{Cov}(q_m)}{\sum_{\mathcal{N}_c(q_{i, k})}  1 - \frac{||q_{i, k} - q_m||}{r_{\text{max}}}},
                \end{align}

                where~$\mathcal{N}_c(q_{i, k})$ defines the $N_c$ nearest covariances $\text{Cov}(q_m)$ (stored from the covariance experiment) to the relative position $q_{i, k}$. The value of $N_c$ is a design parameter with larger $N_c$ giving smoother interpolation at the cost of longer processing time. $r_{\text{max}}$ denotes the L2 Norm of the longest vector that can be made in the FOV of the sensor. To find the detection covariance map required for each iteration of the agent, each agent only stores the covariance for the discrete relative range-pan-tilt triplets obtained from the experiment. This makes the implementation very memory efficient. For the actual experiment, we use $N_c = 1$ as a there is very little processing involved in that case. This yields a Voronoi tessellation of the Sensor FOV for covariance as shown in Figure \ref{fig:VisionCovarianceMap}. We plan to experiment with varying $N$ for future experiments.
                
                %See Appendix~\ref{se:detectiondetails} for details on how to experimentally compute it.       
       
        \subsection{Relative Position Sensing}
        \label{se:relativeposition_hardware}

            Recently, Ultra-WideBand (UWB) based localization systems have been cropping up in popularity and allows precise and accurate localization in indoor and outdoor environments. Most of the work in this system has been done based on Time of Arrival~(ToA)~\cite{Guler2021,Li2020} and Time Difference of Arrival~(TDoA)~\cite{Barral2019} based algorithms, where, between two participating UWB sensors a series of transactions yield the distance between the sensors. However, a lesser explored area is the Angle of Arrival (AoA) estimate using multi antenna array. We have seen some work in this area~\cite{Tiemann2020,Heydariaan2020}, however, as far as the authors are aware, there is no complete 3D implementation of the AoA Scheme using UWB sensors that does not rely on pre-existing tags in the environment. 

            \begin{figure}[h]
                \centering
                \centering
                \includegraphics[width=\linewidth]{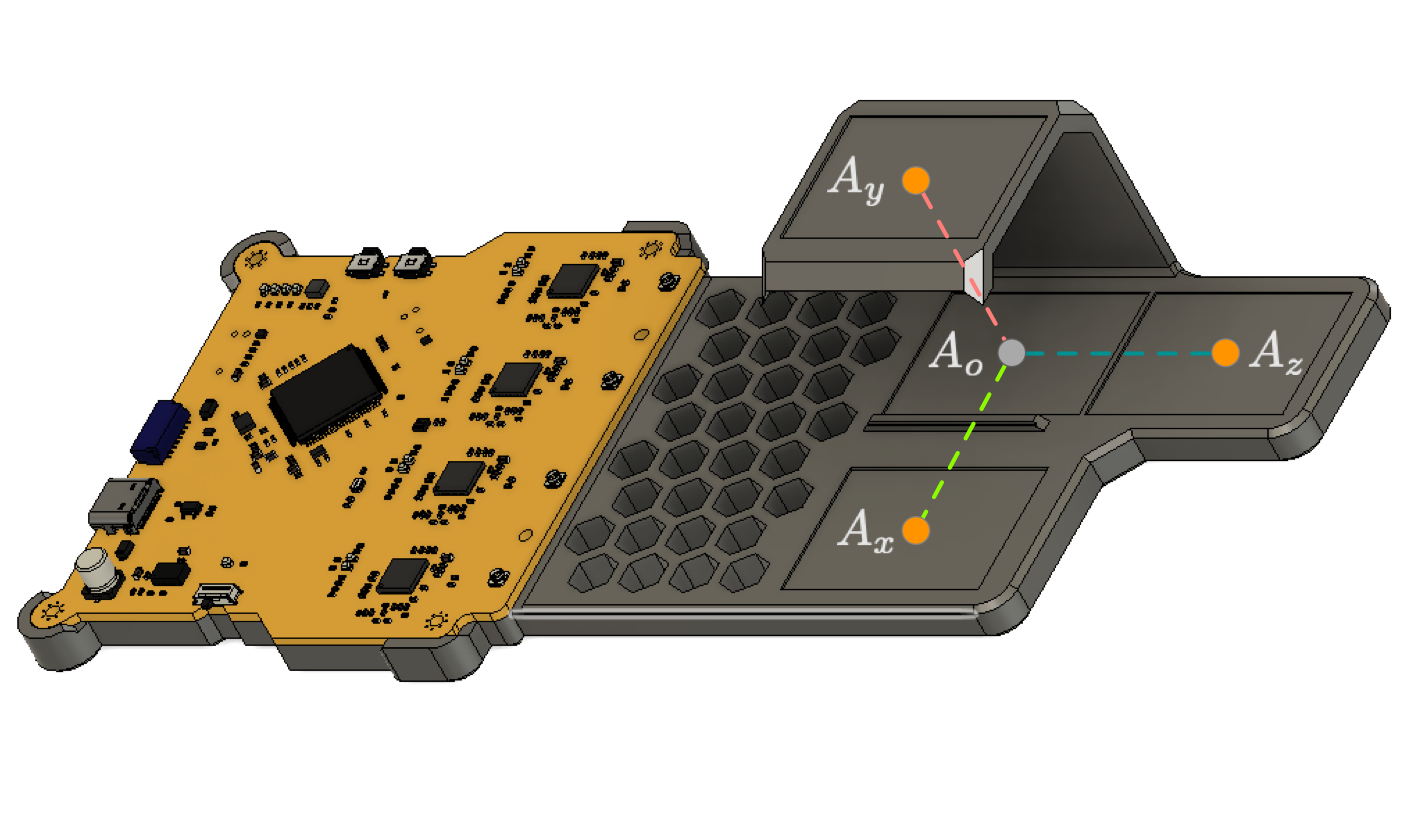}
                \caption{ReLoki Orthogonal Antenna Module. We show the different antenna pairs aligned to the 3 euclidean axes.}
                \label{fig:Reloki_Ortho_Module}
            \end{figure}

            So we developed a novel UWB research platform for full 3D relative localization with no infrastructure requirements that we call ReLoki \cite{mathew2024reloki}. It is a combined communication and localization platform. To put succinctly, the module allows sending a message between agents and simultaneously allowing the receiver node to estimate the relative position~$\hat{p}_{i, j}$ of the transmitter node. We use DW1000 UWB modules for this operation. Each ReLoki transceiver system has 4 DW1000 modules as that is the minimum number of antenna elements required for full 3D localization~\cite{Luo2022,Phalak2020}. The antennas are arranged in an orthogonal array as shown in Figure~\ref{fig:Reloki_Ortho_Module}. To perform localization, 
            the module uses Two way ranging~\cite{Yi2007} for range measurement and AoA measurement for bearing estimate~\cite{Dotlic2018} thereby obtaining the combined relative position of the neighboring agent. 
            The ReLoki module uses a specific message passing and localization protocol as detailed in~\cite{mathew2024reloki}. All the measurements are processed in an onboard processing system. This makes ReLoki a plug and play module for any robotic system. 

            \subsubsection{Covariance Calibration of Relative Position Sensing}\label{se:relativepositiondetails}

                \begin{figure}[ht]
                    \centering
                    \includegraphics[width=0.8\linewidth]{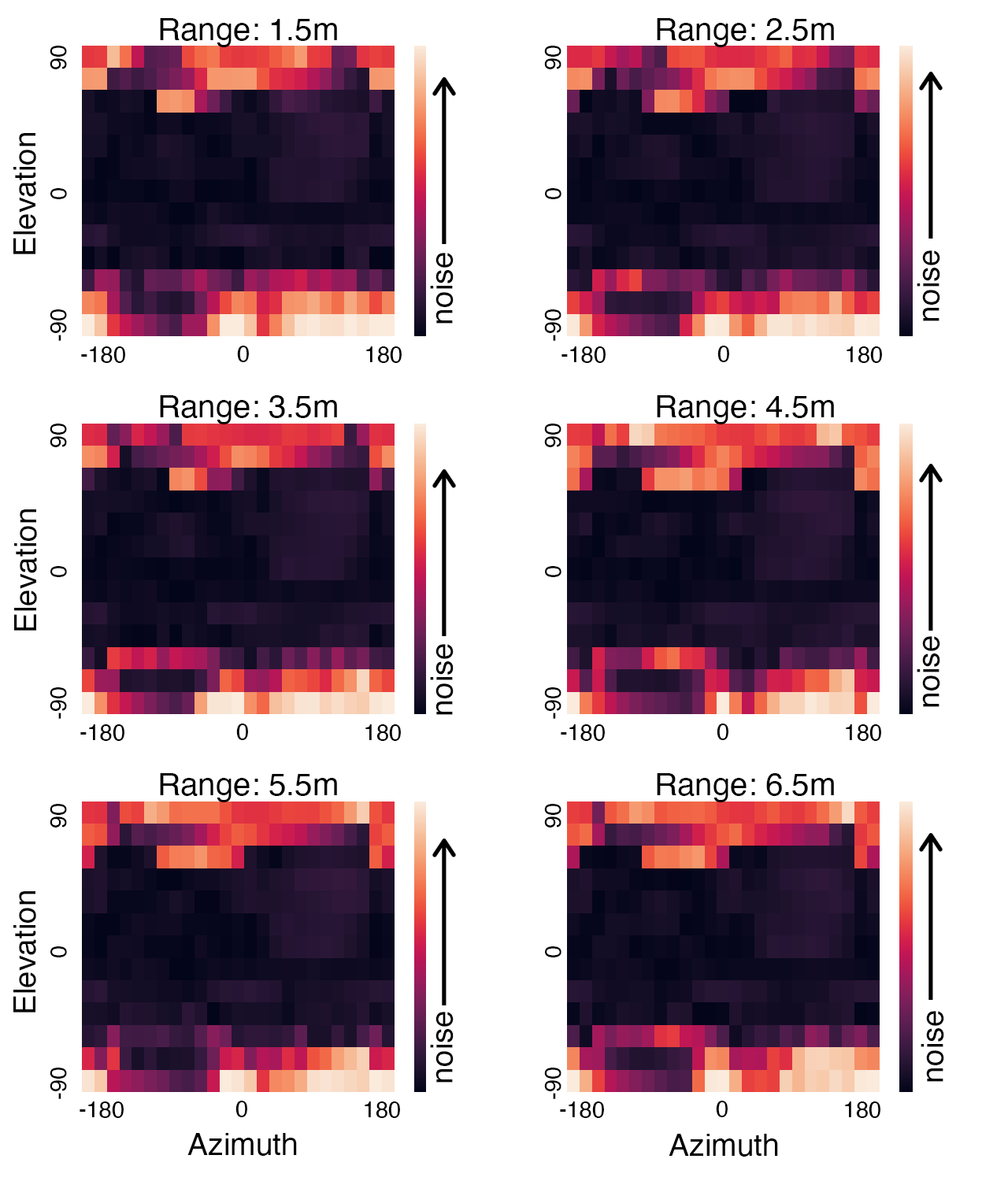}
                    \caption{Entropy of relative comm based relative position sensing using ReLoki. Each image shows the entropy at different bearing angles for a give range}
                    \label{fig:ReLokiCovarianceMap}
                \end{figure}
    
                As was the case with the target detection system, we also found the covariance map for the relative localization system. This is again an experimental setup consisting of the ReLoki acting as the transmitter mounted statically, and the receiver mounted on a pan and tilt joint. The experiment consists of taking 50 readings of relative localization readings at all pan ranges of ~$[-180^\circ, 180^\circ]$ and tilt ranges of ~$[-90^\circ, 90^\circ]$ in steps of~$10^\circ$ and ranges in steps of $1\unit{m}$, starting from $1.5\unit{m}$ with a max range of $6.5\unit{m}$ We find the covariance map using the same procedure used for target detection system. The raw plots of entropy of sensing are shown in Figure~\ref{fig:ReLokiCovarianceMap}. We observe that the detection performance is very good close to $0^\circ$ elevation angle, and it's worse for elevation angles above $60^\circ$. We believe this can be attributed to the use of only one antenna pair for the elevation angle calculation. This can be improved using another antenna configuration which we plan to explore in the future.
    
                We use the same covariance smoothing method explained in the target detection system to get the final covariance used in the proposed algorithm.

        \subsection{Displacement Sensing}\label{se:displacement_hardware}
            To make sure we can do proper feedback control and update the pheromone and target locations relative to the body frame, we need to get the displacement information $\Delta p_i$, showing the movement of the agent since the last update. For displacement sensing we have seen a lot of works already available in the literature. We are using a simple IMU based dead-reckoning (BNO085 IMU) which outputs linear acceleration and orientation of the agent. There is a pre-existing Kalman filter on board the sensor that fuses the accelerometer, gyroscope and magnetometer data. For relative rotational displacement, we use the difference between orientation readings observed at the appropriate time-steps $\Delta p^r_i(t) = p^r_i(t) - p^r_i(t-1)$. For relative translational displacement, the linear acceleration measurements from the IMU is integrated to get a very noisy estimate of the flow data from IMU $\Delta p^{t}_i$. Hence, the displacement output will be $\Delta p_i = [\Delta p^{t}_i, \Delta p^{r}_i]$.

            \subsubsection{Displacement sensing Covariance Map}\label{se:displacementSensingDetails}

                Due to the very noisy nature of the IMU readings, we have observed that the covariance for this system $R_i^{\Delta p}(.)$ is almost uniform over the entire domain. So we approximate the covariance map with a single value: 
                    
                \begin{align*}
                R^{\Delta p}_{i}(.) = \text{diag}[0.18451235 \quad 0.20685948 \quad 0.10835263  \\ 0.000181012 \quad 0.000183857 \quad 0.000290238]
                    \end{align*}
            
                \end{appendices}

\end{document}